\documentclass[lettersize,journal]{IEEEtran}
\usepackage{amsmath,amsfonts}
\usepackage{algorithmic}
\usepackage{array}
\usepackage{textcomp}
\usepackage{stfloats}
\usepackage{url}
\usepackage{bm}
\usepackage{verbatim}
\def\BibTeX{{\rm B\kern-.05em{\sc i\kern-.025em b}\kern-.08em
    T\kern-.1667em\lower.7ex\hbox{E}\kern-.125emX}}
\usepackage{balance}
\usepackage{diagbox}
\usepackage[ruled,vlined]{algorithm2e}
\usepackage{tikz}
\usepackage{enumitem}
\usepackage{balance}
\usepackage{booktabs} 
\usepackage{graphicx}
\usepackage{url}
\usepackage{tikz}
\usepackage{adjustbox}
\usepackage{booktabs}
\usepackage{rotating}
\usepackage{multirow}
\usepackage{lineno}
\usepackage{color, colortbl}
\definecolor{grey}{gray}{0.9}
\usepackage{hyperref}
\hypersetup{colorlinks,linkcolor={blue},citecolor={blue},urlcolor={blue}}  
\usepackage[misc,geometry]{ifsym}
\usepackage{dirtytalk}
\usepackage[normalem]{ulem}
\usepackage{blindtext}
\usepackage{tabularx}
\usepackage{tabulary}
\usepackage{wrapfig}
\usepackage{amssymb}
\usepackage{multirow}
\usepackage{float}
\usepackage{svg}
\usepackage{cite}
\usepackage{physics}
\usepackage[cmintegrals]{newtxmath}
\begin{document}

\date{}

\title{\huge An Adversarial Attack Analysis on Malicious Advertisement URL Detection Framework}

\author{Ehsan Nowroozi,~\IEEEmembership{Member,~IEEE,} Abhishek,~\IEEEmembership{Member,~IEEE,} Mohammad~Reza~Mohammadi,~\IEEEmembership{Member,~IEEE}, and Mauro Conti, ~\IEEEmembership{Fellow Member,~IEEE}
\thanks{E. Nowroozi is a Postdoctoral researcher in Faculty of Engineering and Natural Sciences (FENS), Center of Excellence in Data Analytics (VERİM), Sabanci University, Istanbul Turkey 34956. He is a also external collaborator at the University of Padua, Security and Privacy Research Group (SPRITZ), Italy. (e-mail:ehsan.nowroozi65@gmail.com, ehsan.nowroozi@sabanciuniv.edu, nowroozi@math.unipd.it).}
\thanks{Abhishek is a researcher affiliated from Department of Mathematics, MNNIT Allahabad, India and affiliated from Mathematics Department, University of Delhi-110007, India (e-mail:abhishek1verma99@gmail.com).}
\thanks{M. R. Mohammadi, is with the Department of Mathematics, Security and Privacy Research Group (SPRITZ), University of Padua, 35121, Padua, Italy. (email: mohammadreza.mohammadi@studenti.unipd.it).}%
\thanks{M. Conti is a professor in the Department of Mathematics and the Director of the Security and Privacy Research Group (SPRITZ) at the University of Padua, Italy.  (e-mail: mauro.conti@unipd.it).}}


\maketitle

\begin{abstract}
Malicious advertisement URLs pose a security risk since they are the source of cyber-attacks, and the need to address this issue is growing in both industry and academia. Generally, the attacker delivers an attack vector to the user by means of an email, an advertisement link or any other means of communication and directs them to a malicious website to steal sensitive information and to defraud them. Existing malicious URL detection techniques are limited and to handle unseen features as well as generalize to test data.

In this study, we extract a novel set of lexical and web-scrapped features and employ machine learning technique to set up system for fraudulent advertisement URLs detection. The combination set of six different kinds of features precisely overcome the obfuscation in fraudulent URL classification. Based on different statistical properties, we use twelve different formatted datasets for detection, prediction and classification task. We extend our prediction analysis for mismatched and unlabelled datasets. For this framework, we analyze the performance of four machine learning techniques: Random Forest, Gradient Boost, XGBoost and AdaBoost in the detection part. With our proposed method, we can achieve a false negative rate as low as 0.0037 while maintaining high accuracy of 99.63\%. Moreover, we devise a novel unsupervised technique for data clustering using K-Means algorithm for the visual analysis. This paper analyses the vulnerability of decision tree-based models using the limited knowledge attack scenario. We considered the exploratory attack and implemented Zeroth Order Optimization adversarial attack on the detection models.
\end{abstract}


\section{Introduction}
\IEEEPARstart{I}{n} recent years, the network of web pages has grown faster with the expansion of the Internet. Online services, business, banking, and online marketing have made the Internet an integral part of our lives. Because of the numerous advantages of this platform for online advertising, it has also become a primary source of malicious activities. Attackers deliberately put malicious links in online advertisements that, when visited, redirect the users to unauthorised websites. Attackers make it easy for people to be steered to phishing or malware websites to steal their confidential data, make a fast buck, or defraud them by injecting dangerous code into these websites. Every year, such illegal activities cost billions of dollars \cite{ref3}. Most harmful websites are almost identical to genuine websites, and the user cannot distinguish between them. In order to minimise the effects of this scam, organisations and enterprises are investing a significant amount of money in keeping their systems secure against these harmful links and URLs.

The researchers made several attempts to distinguish the malicious URL using statistical analysis and the popularity feature of the domain name  \cite{ref1}. Though most phishing and malicious sites have a short lifespan, these features may not be available to them for analysis. As a result, the rate of misclassification rises. The creation of new URLs daily makes the job more challenging. URL analysis also serves as a barrier between the victim and the malicious website. There are currently two primary trends in identifying malicious URLs: the first is detecting malicious URLs based on rule's set, and the second is based on a behavioural analysis approach. The approach based on rules set can efficiently and precisely detect harmful URLs. However, this strategy cannot recognise latest malicious advertisement URLs which do not match any of the specified indicators. Deep learning and machine learning algorithms have wide use in categorising malicious URLs based on their behaviour analysis techniques. Classification tasks across a wide range of domains have been found to perform better with deep learning approaches than feature-based learning approaches. Nevertheless, the performance of every deep learning model is determined by a variety of aspects such as hyper-parameter tuning, neural network design, and more.
\subsection{\textbf{Contribution}}
The following is a summary of our contributions to this article:
\begin{itemize}
    \item We devise and integrate the collection of innovative lexical and web-scrapped features \cite{ref4} for the precise categorisation of legitimate and malicious advertisement URLs. Additionally, we utilize several essential human-engineered and URL segmentation features. Furthermore, we examine a range of statistical properties of the URL for a brief idea about the malicious and legitimate class of URL.
    \item We conduct our study on a number of datasets with various URL schemes and orientations, and we analysed our model's performance on these datasets. This method aims to train machine learning models for all URL forms thoroughly.
    \item We investigate the performance of our four distinct classifiers using a range of parameters and the detection results using a balanced matched and mismatched URL dataset. With the right combination of features, we achieve the highest detection accuracy up to 99.63\% and the lowest false negative rate as low as 0.0037 in detection.
    \item We use the K-Means clustering technique to integrate the result in a visual comprehension with the formation of clusters. We also make use of the elbow approach to determine the number of cluster classes.
    \item We analyse how adversarial attacks affect our ensembles and our classifiers' vulnerability. We use the Zeroth Order Optimization black-box approach on our models to account for the limited knowledge scenario.
\end{itemize}

\subsection{\textbf{Organisation}}
This paper is divided into six sections that begin with the Introduction section, afterwards the related work in Section \ref{Related_Work}, Background task and knowledge of our framework in Section \ref{Background}, followed by the feature extraction process, machine learning-based malicious URL detection model, prediction, clustering, and adversarial attack implementation in Section \ref{Methodology}, then presenting the outcomes of our approach, as well as a comparative study of past work's results, and discussion in Section \ref{Results_and_Discussion}, and lastly, we discussed the conclusion of our research and future work in Section \ref{CONCLUSION_AND_FUTURE WORK}. Our code and all the datasets are available on GitHub \cite{ref49}. 
\section{Background}
\label{Background}
Here, we lay out the background information of datasets used in our experiment in Section \ref{datasets}, statistical properties in Section \ref{statisticalcharacteristicsofthedatasets}, and pre-processing task in Section \ref{preprocessingofdatasets}. Here, we will also talk about classification methods in Section \ref{classificationtechniques} and the setting of an adversarial attack in Section \ref{adversarialattack}.
\subsection{\bf\emph{Datasets}}
\label{datasets}
The experiment setup for advertising URLs from $12$ distinct datasets includes $3980870$ URLs. There are two kinds of URLs in these contained in these datasets: benign and malicious. Furthermore, the malicious URL dataset includes four distinct sub-categories: spam, defacement, malware, and phishing. We also examined all of the URLs using the VirusTotal \cite{ref47} tool to confirm their authenticity. Each URL is labelled with '0' for benign and a '1' for malicious in the dataset. The URLs that performed the classification task are divided into two categories listed here.

\subsubsection{\bf\emph{Benign Dataset}}
The benign URLs were gathered from available sources on the Internet. Table \ref{table:benigndatasets} gives an overview of the six benign URL datasets. In Alexa, \cite{ref38} traffic rating is used to rate the URLs, which is calculated by a combination of the browsing behaviour of online users, the total unique visitors, and the number of site traffic. CrowdFlower \cite{ref39} is a large-enhanced dataset of categorised websites; contributors visited supplied links and chose a primary and sub-category for URLs. DMOZ \cite{ref40} is a big open directory that is communally managed and categorises web material. The information is presented in a sophisticated XML format. Benign set URL\cite{ref41} and Non-malicious URL\cite{ref42} are the open-source datasets available on Kaggle. ISCX-URL-2016\cite{ref43} is produced by the research and development unit Canadian Institute for Cybersecurity that draws on the experience of social researchers, business researchers, computer scientists, engineers, lawyers, and scientists. \cite{ref36}. If we talk about the format, Alexa and CrowdFlower datasets have sub-domain and domain in the URL, DMOZ and Benign set URLs (Benign) have protocol and hostname, and Non-malicious\_url and ISCX-URL-2016 (Benign) have all URL components. 
\vspace{-5pt}
\begin{table}[h]
\centering
\caption{Benign Datasets}{}
\label{table:benigndatasets}
\begin{tabular}{|p{0.9cm}|p{4cm}|r|p{1cm}|}
\hline
\textbf{Sr. No.} & \textbf{Dataset Name} & \multicolumn{1}{l|}{\textbf{Volume}} & \multicolumn{1}{l|}{\textbf{Year}} \\ \hline \hline
1                & Alexa\cite{ref38}                         & 1000000                              & 2019                \\ 
\hline
2                & CrowdFlower\cite{ref39}                   & 31084                                & 2020                                            \\ \hline
3                & DMOZ\cite{ref40}                          & 1048576                              & 2019                                            \\ \hline
4                & Benign set URL\cite{ref41}                & 345737                               & 2019                                            \\ \hline
5                & Non-malicious URL\cite{ref42}             & 344821                               & 2019                                            \\ \hline
6                & ISCX-URL-2016\cite{ref43}                 & 35378                                & 2016                                            \\ \hline
\end{tabular}
\end{table}

\subsubsection{\bf \emph{Malicious Dataset}}
We gather the malicious URL datasets from various open sources on the Internet and divide into four sub-categories: spam, defacement, phishing, and malware. Table \ref{table:maliciousdatasets} contains the summary of the six malicious URL datasets. As an open community website, Phishtank\cite{ref45} is a free service for sharing phishing URLs, and users can send suspicious URLs to Phishtank for verification. Cisco Talos Intelligence Group updates data on this website on an hourly basis. Phishstrom\cite{ref46} is a malicious dataset with features built and used for evaluation in the paper \cite{ref37}. Phishing Site URL\cite{ref44}, Malicious data URL\cite{ref42}, and Malicious set URl\cite{ref41} are the open-source datasets available on the web. ISCX-URL-2016 (malicious)\cite{ref43} dataset of malicious URLs contains four different sub-categories of malicious URLs. All of the malicious datasets in this section have complete URL components.

\vspace{-5pt}
\begin{table}[!ht]
\centering
\caption{Malicious Datasets}
\label{table:maliciousdatasets}
\begin{tabular}{|p{0.9cm}|p{4cm}|r|p{1cm}|}
\hline

\textbf{Sr. No.} & \textbf{Dataset Name}          & \multicolumn{1}{l|}{\textbf{Volume}} & \multicolumn{1}{l|}{\textbf{Year}} \\ \hline \hline
1                & Phishing Site URL\cite{ref44}              & 549347                               & 2020                                            \\ \hline
2                & Phishtank\cite{ref45}                      & 223056                               & 2021                                            \\ \hline
3                & Malicious data URL\cite{ref42}             & 75643                                & 2019                                            \\ \hline
4                & ISCX-URL-2016\cite{ref43}                  & 129988                               & 2016                                            \\ \hline
5                & Phishstrom\cite{ref46}                     & 47909                                & 2013                                            \\ \hline
6                & Malicious set URL\cite{ref41}              & 104438                               & 2019                                            \\ \hline
\end{tabular}
\end{table}

The six benign and six malicious URL datasets are combined to create six balanced datasets which contain an equal number of malicious and benign URLs with a split size of half of the benign data. The benign datasets are then combined based on their resemblance to malicious datasets. The combined information from the merged datasets is also displayed in Table \ref{table:mergeddatasets}.

\begin{table}[!ht]
\centering
\caption{Merged Datasets}
\label{table:mergeddatasets}
\begin{tabular}{|c|l|}
\hline 
\multicolumn{1}{|l|}{\textbf{Dataset Name}} & \textbf{Merged Dataset}                                                                                   \\ \hline \hline
Dataset 1                                   & \begin{tabular}[c]{@{}c@{}}Alexa + Phishing Site URL\end{tabular}              \\ \hline
Dataset 2                                   & \begin{tabular}[c]{@{}c@{}}Crowdflower + Phishtank\end{tabular}                  \\ \hline
Dataset 3  & \begin{tabular}[c]{@{}c@{}}DMOZ + Malicious data URL \end{tabular}   \\ \hline
Dataset 4                                   & \begin{tabular}[c]{@{}c@{}}Benign set URL + ISCX-URL-2016 (Malicious)\end{tabular}  \\ \hline
Dataset 5                                   & \begin{tabular}[c]{@{}c@{}}Non-malicious URL + Phishstrom \end{tabular}      \\ \hline
Dataset 6                                   & \begin{tabular}[c]{@{}c@{}}ISCX-URL-2016 (Benign) + Malicious set URL \end{tabular} \\ \hline
\end{tabular}
\end{table}

\subsection{\bf \emph{Statistical Characteristics of the Datasets}}
\label{statisticalcharacteristicsofthedatasets}
\begin{figure}[!ht]
\centering
\includegraphics[width=3.5in]{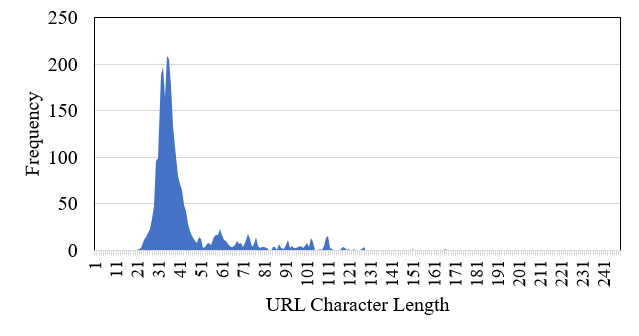}
\caption{Frequency Distribution of URL Length - Benign}
\label{fig:frequencydistributionofurllengthbenign}
\end{figure}

To better understand our data, we perform a statistical study of legitimate and malicious advertisement URLs based on lexical properties here. In this analysis, we discovered that the length of malicious ad URLs has a broader range of dispersion than benign URLs. The majority of benign URLs are between $25$ and $50$ characters long, with an average length of $44.28$ characters. The average number of special characters in a benign URL is $8.64$, with a path length of $17.54$ characters. Fig. \ref{fig:frequencydistributionofurllengthbenign} shows the frequency distribution of benign URL lengths.

\begin{figure}[!ht]
\centering
\includegraphics[width=3.5in]{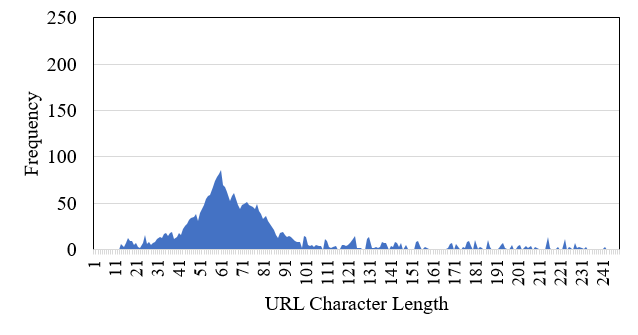}
\caption{Frequency Distribution of URL Length - Malicious}
\label{fig:frequencydistributionofurllengthmalicious}
\end{figure}

The majority of malicious ad URLs are between $25$ and $105$ characters long, with an average length of $63.14$ characters. Malicious URLs have $13.98$ special characters on average and a path length of $42.60$ characters. We also discovered that less than $2$ per cent of malicious ad URLs use IP as a domain name. Fig. \ref{fig:frequencydistributionofurllengthmalicious} presents the frequency distribution of malicious URLs.

\subsection{\bf \emph{Preprocessing of Datasets}}
\label{preprocessingofdatasets}
Our prototype platform's implementation accepts a specified shape of the input dataset. As a result, to acquire more precise outcomes, the dataset must be modified and reshaped to meet our reliance's required input format. We scale the data using the interquartile range method during the preprocessing stage. All redundant and raw valued cells were removed, and repeated hostname URLs were excluded. The URL datasets were then shuffled, and samples were taken from the datasets for further investigation.

\subsection{\bf\emph{Classification Techniques}}
\label{classificationtechniques}
The Classification method is a Supervised Learning technique that uses training data to identify the category of new observations. Predictive models train from a given dataset or collection of observations and afterwards categorize subsequent observations into one of many classes or groupings. Classes are sometimes known as targets/labels or categories. Boosting is an ensemble modelling approach that aims to construct a robust classifier from many weak classifiers. This task is performed by building a model by sequentially connecting weak models. First, a model is constructed using the training data. Then the second model is constructed to address the faults in the previous model. The is repeated again and again until the entire training data set is adequately predicted or the maximum number of models added. Here, we discussed some of the classifications and boosting techniques we used in our research.

\subsubsection{\bf \emph {Random Forest}}
A Random Forest is a large collection of decision trees that differ slightly from the others. Random Forest is based on the premise that while every tree may predict quite well, it almost certainly overfits some data. Overfitting can be limited by averaging the outcomes of numerous trees that all operate well and overfit in different ways. This reduction in overfitting can also be demonstrated while keeping the predictive power of the trees. The important parameter to alter are \textit{n\_estimators}, \textit{max features}, and pre-pruning settings such as \textit{max depth}. For \textit{n\_estimators}, larger value is preferred. As more trees are averaged, an ensemble becomes more robust as overfitting is reduced. However, more trees require more memory and training time. 

\subsubsection{\bf \emph {Gradient Boost}}
Gradient boosting is another ensemble approach that combines numerous decision trees to generate a more significant model. Unlike the Random Forest method, Gradient Boosting works by successively creating trees for every tree, attempting to fix the flaws of the preceding one. Randomization is not employed by default in Gradient Boost regression trees; instead, severe pre-pruning is used. Gradient Boost trees usually employ extremely shallow trees with depths ranging from one to five, reducing memory needs and speeding up the prediction. Gradient Boost tree models depend on three primary parameters: \textit{n\_estimators}, \textit{number of trees}, and \textit{learning rate}, that deduce the extent by which new tree corrects the errors of the previous one. It is common practice to fit \textit{n\_estimators} for learning rates according to a time and memory budget and then comparing.

\subsubsection{\bf \emph {AdaBoost}}
Another ensemble method in Machine Learning, AdaBoost, or Adaptive Boosting, is a boosting algorithm. During the data training period, it generates \textit{n} decision trees. When the first decision tree/model is constructed, the incorrectly classified record in the first model is prioritised. Only these records are sent to the second model as input. The process is repeated until we specify how many base learners we want to create. With all boosting techniques, record repetition is permitted. The incorrectly classified record is used as input for the next model. This process is repeated until the specified condition is met.

\subsubsection{\bf \emph {XGBoost}}
XGBoost, or extreme Gradient Boosting, enhances performance and speed aspects of Gradient Boost decision trees implementation. To create a machine learning algorithm, it used a Gradient Boost framework. The regularisation term reduces over-fitting by smoothing the final learned weights. Models using basic, predictive functions will be favoured by the regularised objective. Here, the model is taught additively instead of the traditional optimization method for the tree ensemble model in the Euclidean space. Due to its simultaneous and distributed processing, XGBoost is faster than other algorithms.

\subsection{\bf \emph{Adversarial Attack}}
\label{adversarialattack}
Adversarial machine learning technique aims to exploit models by developing harmful attacks using accessible model information \cite{IEEEtnsm2022}. The main reason for exploitation is to affect machine learning and fail it. Here, we will examine the adversarial attack on our models. It has been discovered that an intelligent malicious URL generating system can generate URLs that can pass machine learning classifiers at a low rate. 
White-box, gray-box, and black-box attacks are the three main types of adversarial attacks.
\begin{itemize}
\item{{\bf \emph {White-Box Attack}}}: In a white-box attack, the attackers have complete access to the model's architecture, parameters, weights. L-BFGS \cite{ref53} black-box constrained method was proposed to minimise the additive perturbation based on classification restrictions; however, it was inefficient. The Fast Gradient Sign Method \cite{ref48} eventually conquered it, followed by the primary iterative method of adding perturbations iteratively. Other white-box adversarial attacks include JSMA, BIM \& ILCM, ATNs, and DeepFool. 
\item{{\bf \emph {Black-Box Attack}}}: In black-box attacks, the attacker have no access to the system. The system's parameters and weights cannot be obtained here, opposite to white-box attack, which need the target neural networks to be differentiable. Zeroth Order Optimization may directly approximate the gradients of the target network using zeroth order stochastic coordinate descent. Other examples of black-box adversarial attacks include One-pixel, UPSET, and ANGRI. 
%
\item{{\bf \emph {Grey-Box Attack}}}: The grey-box attacks use only training access to the target model to create adversarial samples.\\
\end{itemize}

\textbf{Counter-Forensic Attack Model:}
Here we shall formulate the universal adversarial paradigm for counter forensic attacks in this text. An adversarial model is characterised by demonstrating adversary's goal, system's knowledge, and capacity of corrupting the system through data manipulation\cite{ref54, nowroozi2020machine}.
\begin{itemize}
\item {\textit{Attacker's goal:}} It describes the type of security violation and the type of error the attacker seeks. Counter forensic attacks are often either integrity violations or evasion attacks. The adversary’s goal determines the loss function that the adversary seeks to maximise. 

\item \textit{Attacker's knowledge:} This deals with if the attack has a limited amount of knowledge about the model or perfect knowledge. The attacker is fully aware of the forensic algorithm in a perfect knowledge scenario and aware of a few details or settings relevant to the forensic algorithms in a restricted knowledge scenario.

\item \textit{Attackers capability:} It refers to the adversary's level of control over training and testing data. The attacker's capability in the exploratory attack paradigm is restricted to modifications to test data, whereas modifications to training examples are still not permissible. The attack can interrupt the training process in a causative attack scenario, referred to as a poisoning attack. 
\end{itemize}

This article considers the exploratory attack scenario and shall discuss the ZOO attack. The ZOO attack is a version of the C\&W attack\cite{ref21} that employs ADAM coordinate descent to perform numerical gradient estimates. As with any tree attack, there is the risk of misclassification of objects or data. It can also involve picture captioning, voice recognition misclassification, and data regression effects. Because a white-box attack relies on the model's overall knowledge growth, this black box attack has just a limited knowledge of the targeted model. It has an unknown training technique as well as data. It also includes unidentified output classes and model confidence.

Many research challenges are concerned with complex data creation processes that, although analytically explainable, can provide types of measurements such as measures from physical surroundings or predictions from deploying machine learning techniques. In black-box models, these kinds of challenges are incorporated into Zeroth Order Optimization (ZOO).These sorts of issues are put into Zeroth Order Optimization in black-box models. The ZOO Attack, also known as Derivative-Free Optimization, solves problems that do not have direct access to gradients.

\begin{table*}
\centering
\caption{Lexical Feature Group}
\label{table:lexicalfeaturegroup}
\begin{tabular}{|l|l|l|l|l|} 
\hline
\textbf{Sr. No.} & \textbf{Feature Subgroup}                                       & \textbf{Feature Name}    & \textbf{Data Type} & \textbf{Feature Description}                                                                                                               \\ 
\hline \hline
1                & \multirow{40}{*}{\textbf{Linguistic Features}}                                                & URLLength                & Integer            & Compute length of the URL e.g. the URL length is 51 of \\
& & & & https://www.example.com/seo-tools/count-characters/                               \\ 
\cline{1-1}\cline{3-5}
2                &                                                                                      & CheckIPAsHostName        & Binary             & Check if IP address is used as hostname e.g. www.192.168.0.1                                                     \\ 
\cline{1-1}\cline{3-5}
3                &                                                                                      & CheckEXE                 & Binary             & To look the presence of .exe in URL                                                                                             \\ 
\cline{1-1}\cline{3-5}
4                &                                                                                      & DigitAlphabetRatio       & Float              & To compute the ratio of number of digit to alphabets in URL                                                              \\ 
\cline{1-1}\cline{3-5}
5                &                                                                                      & SpecialcharAlphabetRatio & Float              & To compute the ratio of number of special characters to alphabets in URL                                                 \\ 
\cline{1-1}\cline{3-5}
6                &                                                                                      & UppercaseLowercaseRatio  & Float              & It computes ratio of uppercase characters to lowercase characters in URL                                    \\ 
\cline{1-1}\cline{3-5}
7                &                                                                                      & DomainURLRatio           & Float              & To compute the ratio of domain length to URL length                                                                                \\ 
\cline{1-1}\cline{3-5}
8                &                                                                                      & NumericCharCount         & Integer            & Number of numeric character viz. 0,1,2,...,9                                                                                       \\ 
\cline{1-1}\cline{3-5}
9                &                                                                                      & EnglishLetterCount       & Integer            & Number of English letter viz. a,b,c,...,z and A,B,C,...,Z                                                                          \\ 
\cline{1-1}\cline{3-5}
10               &                                                                                      & SpecialCharCount         & Integer            & Number of special character viz. !,\$,”,...etc.                                                                                    \\ 
\cline{1-1}\cline{3-5}
11               &                                                                                      & DotCount                 & Integer            & Number of ‘.’ characters                                                                                                        \\ 
\cline{1-1}\cline{3-5}
12               &                                                                                      & SemiColCount             & Integer            & Number of ‘;’ characters                                                                                                        \\ 
\cline{1-1}\cline{3-5}
13               &                                                                                      & UnderscoreCount          & Integer            & Number of ‘\_’ characters                                                                                                       \\ 
\cline{1-1}\cline{3-5}
14               &                                                                                      & QuesMarkCount            & Integer            & Number of ‘?’ characters                                                                                                        \\ 
\cline{1-1}\cline{3-5}
15               &                                                                                      & HashCharCount            & Integer            & Number of ‘\#’ characters                                                                                                        \\ 
\cline{1-1}\cline{3-5}
16               &                                                                                      & EqualCount               & Integer            & Number of ‘=’ characters                                                                                                         \\ 
\cline{1-1}\cline{3-5}
17               &                                                                                      & PercentCharCount         & Integer            & Number of ‘\%’ characters                                                                                                        \\ 
\cline{1-1}\cline{3-5}
18               &                                                                                      & AmpersandCount           & Integer            & Number of ‘\&’ characters                                                                                                          \\ 
\cline{1-1}\cline{3-5}
19               &                                                                                      & DashCharCount            & Integer            & Number of ‘-’ characters                                                                                                         \\ 
\cline{1-1}\cline{3-5}
20               &                                                                                      & DelimiterCount           & Integer            & Number of the dilimeter characters in URL viz. (),\{\}{[}],,/*,*/,...etc.                                                           \\ 
\cline{1-1}\cline{3-5}
21               &                                                                                      & AtCharCount              & Integer            & Number of '@' characters                                                                                                  \\ 
\cline{1-1}\cline{3-5}
22               &                                                                                      & TildeCharCount           & Integer            & Number of ‘$\sim$’ characters                                                                                            \\ 
\cline{1-1}\cline{3-5}
23               &                                                                                      & DoubleSlashCount         & Integer            & Number of ‘//’ characters in the link path                                                                                      \\ 
\cline{1-1}\cline{3-5}
24               &                                                                                      & IsHashed                 & Binary             & To check if URL is hashed (If any hash function is used to convert URL)                                              \\ 
\cline{1-1}\cline{3-5}
25               &                                                                                      & TLD                      & String             & To look for top level domain of URL e.g. .com, .us, .org                                                                                                                   \\ 
\cline{1-1}\cline{3-5}
26               &                                                                                      & DistDigitAlphabet        & Float              & Distance between number to alphabet                                                                                                \\ 
\cline{1-1}\cline{3-5}
27               &                                                                                      & HttpsInUrl               & Binary             & Check the presence of https in URL                                                                                        \\ 
\cline{1-1}\cline{3-5}
28               &                                                                                      & FileExtension            & String             & To check the extension of the file in URL                                                                                          \\ 
\cline{1-1}\cline{3-5}
29               &                                                                                      & TLDInSubdomain           & Binary             & Check whether subdomain have TLD or ccTLD as its part                                                            \\ 
\cline{1-1}\cline{3-5}
30               &                                                                                      & TLDInPath                & Binary             & Check whether subdomain have TLD or ccTLD in link of URL                                                                           \\ 
\cline{1-1}\cline{3-5}
31               &                                                                                      & HttpsInHostName          & Binary             & Check for the disarrangement of https in URL e.g. ‘www.wordforhttps.in’                                             \\ 
\cline{1-1}\cline{3-5}
32               &                                                                                      & HostNameLength           & Integer            & Hostname length of URL                                                                                                                 \\ 
\cline{1-1}\cline{3-5}
33               &                                                                                      & PathLength               & Integer            & Path length of URL                                                                                                           \\ 
\cline{1-1}\cline{3-5}
34               &                                                                                      & QueryLength              & Integer            & Length of the query In URL                                                                                                               \\ 
\cline{1-1}\cline{3-5}
35               &                                                                                      & DistWordBased            & Binary             & Check if URL is contain anonymous words (personal, .bin, abuse etc.)                                               \\ 
\cline{1-1}\cline{3-5}
36               &                                                                                      & URLWithoutwww            & Binary             & To check if www is present in the URL                                                                                              \\ 
\cline{1-1}\cline{3-5}
37               &                                                                                      & FTPUsed                  & Binary             & Check if “ftp://” is there in the URL                                                                                              \\ 
\cline{1-1}\cline{3-5}
38               &                                                                                      & JSUsed                   & Binary             & Check if “.js” is there in the URL                                                                                                 \\ 
\cline{1-1}\cline{3-5}
39               &                                                                                      & FilesInURL               & Binary             & Check if files is there in the URL                                                                                                 \\ 
\cline{1-1}\cline{3-5}
40               &                                                                                      & CSSUsed                  & Binary             & Check if “.css” is used in the URL                                                                                                 \\ 
\hline
41               & \multirow{9}{*}{\begin{tabular}[c]{@{}c@{}}\textbf{Human-Engineered}\\ \textbf{Features}\end{tabular}} & IsDomainEnglishWord      & Binary             & Check if domain name is English language word as per US/UK dictionary                                                     \\ 
\cline{1-1}\cline{3-5}
42               &                                                                                      & IsDomainMeaningful       & Binary             & To check if domain name is English language word and meaningful                                                                    \\ 
\cline{1-1}\cline{3-5}
43               &                                                                                      & IsDomainPronounceable    & Binary             & To check if domain name can be pronounced                                                                                          \\ 
\cline{1-1}\cline{3-5}
44               &                                                                                      & IsDomainRandom           & Binary             & To compute the randomness of the URL string                                                                                        \\ 
\cline{1-1}\cline{3-5}
45               &                                                                                      & Unigram                  & Float              & To calculate uni-gram probability                                                                                                  \\ 
\cline{1-1}\cline{3-5}
46               &                                                                                      & Bigram                   & Float              & To calculate bi-gram probability                                                                                                   \\ 
\cline{1-1}\cline{3-5}
47               &                                                                                      & Trigram                  & Float              & To calculate tri-gram probability                                                                                                  \\ 
\cline{1-1}\cline{3-5}
48               &                                                                                      & SensitiveWordCount       & Integer            & Sensitive words in URL (i.e., “secure”, “account”, “webscr”, “login” etc)  \\ 
\cline{1-1}\cline{3-5}
49               &                                                                                      & InSuspiciousList         & Binary             & To check if URL is present in suspicious list from malicious dataset                                                               \\
\hline
\end{tabular}
\end{table*}

\section{Related Work}
\label{Related_Work}

Significant research has been conducted earlier to detect malicious advertising and phishing websites in a variety of ways. The traditional approach of blacklisting the malicious ads related sites gives out as the list consists of only known malicious URLs \cite{ref10}. Although it is a powerful feature indication, it takes several hours for a malicious URL to be banned, and its implementation is infeasible. As a result, the system fails when a new malicious URL is encountered. Researchers later proposed URL analysis based on lexical aspects \cite{ref9}, statistical properties \cite{ref8}, host-based features \cite{ref29}, content features \cite{ref31}, and popularity-based factors \cite{ref34}. Thomas et al. \cite{ref12} developed a real-time URL spam filtering process that involved based on IPs, lexical and HTML parameters. The popularity features are the details about the URL's position on social media or its rating. Choi et al. \cite{ref13} used URL lexical analysis to identify multiple forms of malicious assaults. Xiang et al. \cite{ref14} employed machine learning algorithms with a large collection of rich features. In Twitter streams, Lee and Kim \cite{ref15} discovered suspicious URLs by examining and analysing associated redirect chains of URLs in several tweets. Furthermore, determining which features are valuable necessitates specialist topic knowledge, yet our strategy of combining the innovative combination of rich features proved to be thoughtful exertion.

Based on the machine learning approaches, a number of fraudulent advertisements URL detection systems were created, including logistic regression \cite{ref50}, Nave Bayes \cite{ref51}, decision trees, and ensembles \cite{ref52}. Garera et al. \cite{ref16} research is based on logistic regression, with a few variables incorporated in the extraction to categorise harmful URLs [16]. Fette et al. \cite{ref17} classify phishing emails using statistical approaches from machine learning. The classifiers examine the structure and content of the email. Bergholz et al. improve on the technique mentioned above's accuracy by incorporating text classification algorithms to examine email content \cite{ref18}. Abu-Nimeh et al. \cite{ref19} compares multiple classifiers on a database of phishing emails, using the frequency of the corpus's top 43 keywords as attributes. Numerous online tools are available on the web that can help detect malicious URLs, such as URL Void, UnMask Parasites,  Norton Safe Web, Google Safe Browsing Diagnostic, SiteAdvisor, VirusTotal, and Browser Defender. However, their effectiveness is limited due to their signature-based nature. In this article, we will look at the most popular ensemble strategies and compare the accuracy of all the models.

Extensive research has been conducted on the adversarial attack on deep learning models \cite{ref21}. An investigation of the robustness of deep neural networks was conducted when they were exposed to small simulated perturbations \cite{ref11}. In this context, the adversarial robustness of machine learning models was also investigated. Barreno et al. work's \cite{ref35} is part of a series of security evaluations of machine learning algorithms that includes various attacks on machine learning techniques and systems. Huan and Wardey-paper Farley's \cite{ref24} laid the groundwork for such work using sophisticated model architecture and parameters, including the use of adversarial examples to improve the classifier's robustness. A reliable technique for identifying adversarial perturbations is required to investigate and assess the robustness of different classifiers to adversarial perturbation.
\section{Methodology}
\label{Methodology}
In this section, we present our working mechanism with feature extraction procedure in Section \ref{featureextraction}, empirical study in Section \ref{empiricalstudy}, and adversarial attack on our detection models in Section \ref{zooadversarialattack}.

\begin{algorithm}[b]
\caption{Check domain is English language word}
\label{alg:checkdomainisenglishlanguageword}
\begin{algorithmic}
\STATE 
\STATE {\textsc{\textbf{input:}}} {domain\_name}
\STATE {\textsc{\textbf{output:}}} {Meaningful, English word, Pronounceable, Random string}
\STATE break domain\_name in words = \textit{words[]}
\STATE \textbf{IF} domain\_name belongs to \textit{wordnet.synsets}
\STATE \hspace{0.5cm}\textit{Return TRUE}
\STATE \textbf{ELSE}
\STATE \hspace{0.5cm}\textit{Return FALSE}
\STATE \textbf{IF} \textit{words[]} is not \textit{NULL}
\STATE \hspace{0.5cm}Check for \textit{parts of speech}
\STATE \textbf{IF} \textit{words[]} has \textit{Noun, Pronoun}
\STATE \hspace{0.5cm}\textit{Return} Meaningful
\STATE \textbf{ELIF} \textit{words[]} has \textit{Verb, Adjective}
\STATE \hspace{0.5cm}\textit{Return} Pronounceable
\STATE \textbf{ELSE}
\STATE \hspace{0.5cm}\textit{Rerurn} Random string
\STATE {\textsc{\textbf{end}}}
\end{algorithmic}
\label{alg1}
\end{algorithm}

\subsection{\textbf{\textit{Feature Extraction}}}
\label{featureextraction}
This phase of research aims to gather important information about the URL string. It entails a set of features with distinguishing characteristics used to specify different dataset categories. Since we wish to differentiate between fraudulent and genuine advertisement URLs, we do this work using two key feature categories: lexical and web-scrapped features. These are covered in the next section.\\

\subsubsection{\textbf{\textit{Lexical Features}}}
We used the fundamental features from the URL string for easy demonstration and some advanced aspects for an unconventional approach in this set of features. These are further classified into two types: linguistic features and human-engineered features. \textbf{Linguistic features} comprise essential URL string qualities such as length-based features (length of the URL string, domain length, etc.), presence of an object (.exe presence in the string, etc.), count-based features (alphabet, special character count, etc.), and TLDs. In this scenario, features such as the digits-to-alphabet ratio are also relevant because most malicious algorithmically created domains have considerably more digits than alphabets. Aside from the bag-of-words features, statistical features help significantly in gaining a quick understanding of the path, query, and URL length.\\ 
If the domain name has any credibility in terms of being a meaningful, easy-to-pronounce, and random string, as proven in Algorithm \ref{alg:checkdomainisenglishlanguageword}, \textbf{Human-Engineered Features} perform well enough in this classification task. We considered the UK/US English dictionary to state the domain name as a meaningful, pronounceable or random string in the identification process. We also collect "n-gram" data for both legitimate and non-legitimate advertising URLs. We generate the bag-of-words by extracting 2-gram and 3-gram tokens from the data mentioned above sources. This allows us to break each input URL into two or three token sequences and verify the presence of the target token in the bag. The sensitive words contained domains have a high chance of being malicious, and we created it as a feature. We also check whether the domain name is present in the list of suspected lists. Please see Table \ref{table:lexicalfeaturegroup} for a complete list of lexical feature groups as well as an overview of the features. 
%
%

\begin{table*}
\centering
\caption{Web-Scrapped Feature Group}
\label{table:webscrappedfeaturegroup}
\begin{tabular}{|l|c|l|l|l|} 
\hline
\textbf{Sr. No.} & \textbf{Feature Subgroup}                                & \textbf{Feature Name} & \textbf{Data Type} & \textbf{Feature Description}                                                                                                                        \\ 
\hline \hline
1                                                          & \multicolumn{1}{c|}{\multirow{13}{*}{\textbf{Deep-Web Features}}} & LevenshteinDistance   & Float              & To calculate Levenshtein distance                                                                                                           \\ 
\cline{1-1}\cline{3-5}
2                                                          & \multicolumn{1}{c|}{}                                    & Entropy               & Float              & To calculate shannon entropy of the URL                                                                                                     \\ 
\cline{1-1}\cline{3-5}
3                                                          & \multicolumn{1}{c|}{}                                    & Hyphenstring          & String             & To get hyphenated domain name of URL for typosquat                                                                                          \\ 
\cline{1-1}\cline{3-5}
4                                                          & \multicolumn{1}{c|}{}                                    & Homoglyph             & String             & To get homoglyph string for typosquat                                                                                                       \\ 
\cline{1-1}\cline{3-5}
5                                                          & \multicolumn{1}{c|}{}                                    & Vowel                 & String             & To get vowel swap string for typosquat                                                                                                      \\ 
\cline{1-1}\cline{3-5}
6                                                          & \multicolumn{1}{c|}{}                                    & Bitsquatting          & String             & To get bitsquatting string for typosquat                                                                                                    \\ 
\cline{1-1}\cline{3-5}
7                                                          & \multicolumn{1}{c|}{}                                    & InsertionString       & String             & To get insertion string for typosquat                                                                                                       \\ 
\cline{1-1}\cline{3-5}
8                                                          & \multicolumn{1}{c|}{}                                    & Omission              & String             & To get omission string for typosquat                                                                                                        \\ 
\cline{1-1}\cline{3-5}
9                                                          & \multicolumn{1}{c|}{}                                    & Repeatition           & String             & To get repitition string for typosquat                                                                                                      \\ 
\cline{1-1}\cline{3-5}
10                                                         & \multicolumn{1}{c|}{}                                    & Replacement           & String             & To get replaced string for typosquat                                                                                                        \\ 
\cline{1-1}\cline{3-5}
11                                                         & \multicolumn{1}{c|}{}                                    & Subdomain             & String             & To get subdomain string for typosquat                                                                                                       \\ 
\cline{1-1}\cline{3-5}
12                                                         & \multicolumn{1}{c|}{}                                    & Transposition         & String             & To get transposition string for typosquat                                                                                                   \\ 
\cline{1-1}\cline{3-5}
13                                                         & \multicolumn{1}{c|}{}                                    & AdditionString        & String             & To get addition string for typosquat                                                                                                        \\ 
\hline
14                                                         & \textbf{URL Segmentation}                                         & GoogleSearchFeature   & Integer            & Array of top 60 results of URL querying the Google search engine                                  \\ 
\hline
15                                                         & \multirow{8}{*}{\textbf{Host-Based Features}}                     & IPAddress             & Numeric            & To get the IP address of the host name                                                                                                      \\ 
\cline{1-1}\cline{3-5}
16                                                         &                                                          & ASNNumber             & Numeric            & To get the ASN number of the URL                                                                                                            \\ 
\cline{1-1}\cline{3-5}
17                                                         &                                                          & ASNCountryCode        & Numeric            & To get ASN country code                                                                                                                     \\ 
\cline{1-1}\cline{3-5}
18                                                         &                                                          & ASN\_CIDR             & Numeric            & To get the CIDR number of the host name                                                                                                     \\ 
\cline{1-1}\cline{3-5}
19                                                         &                                                          & ASNPostalCode         & Numeric            & To get ASN postal code                                                                                                                      \\ 
\cline{1-1}\cline{3-5}
20                                                         &                                                          & ASNCreationDate       & Numeric            & To get the creation date of the domain name                                                                                                 \\ 
\cline{1-1}\cline{3-5}
21                                                         &                                                          & ASNUpdationDate       & Numeric            & To get the updation date of the domain name                                                                                                 \\ 
\cline{1-1}\cline{3-5}
22                                                         &                                                          & DomainAgeInDays       & Numeric            & The duration of a domain's registration 
\\ 
\hline
23                                                         & \multirow{18}{*}{\textbf{Content-Based Features}}                 & ImgCount              & Integer            & To count the images in the webpage                                                                                                    \\ 
\cline{1-1}\cline{3-5}
24                                                         &                                                          & TotalLinks            & Integer            & To count links in the webpage                                                                                                     \\ 
\cline{1-1}\cline{3-5}
25                                                         &                                                          & NumParameters         & Integer            & Number of parameters of the URL                                                                                                             \\ 
\cline{1-1}\cline{3-5}
26                                                         &                                                          & NumFragments          & Integer            & Number of fragments of the URL                                                                                                              \\ 
\cline{1-1}\cline{3-5}
27                                                         &                                                          & BodyTagCount          & Integer            & Counting the body tag in webpage's html source code                                                                                   \\ 
\cline{1-1}\cline{3-5}
28                                                         &                                                          & MetaTagCount          & Integer            & Counting the meta tag in webpage's html source code                                                                                    \\ 
\cline{1-1}\cline{3-5}
29                                                         &                                                          & DivTagCount           & Integer            & Counting the div tag in webpage's html source code                                                                                     \\ 
\cline{1-1}\cline{3-5}
30                                                         &                                                          & FakeLinkInStatusBar   & Binary             & Check whether display of fake URL JS command on MouseOver                                \\ 
\cline{1-1}\cline{3-5}
31                                                         &                                                          & RightClickDisabled    & Binary             & Check the presence of command to disable right click  \\ 
\cline{1-1}\cline{3-5}
32                                                         &                                                          & PopUpWindow           & Binary             & Check the presence of command to start popup window                   \\ 
\cline{1-1}\cline{3-5}
33                                                         &                                                          & CheckMailto           & Binary             & Whether "mailto" is present in HTML source code                                                                                    \\ 
\cline{1-1}\cline{3-5}
34                                                         &                                                          & CheckFrametag         & Binary             & Whether frame or iframe used in HTML source code                                                                                       \\ 
\cline{1-1}\cline{3-5}
35                                                         &                                                          & TitleCheck            & Binary             & In HTML source codes, see whether title tag is empty 
\\ 
\cline{1-1}\cline{3-5}
36                                                         &                                                          & SourceEvalCount       & Integer            & Count of eval () functions in HTML source code                                                                                             \\ 
\cline{1-1}\cline{3-5}
37                                                         &                                                          & SourceEscapeCount     & Integer            & Count of escape () functions in HTML source code                                                                                            \\ 
\cline{1-1}\cline{3-5}
38                                                         &                                                          & SourceExecCount       & Integer            & Count of exec () functions in HTML source code                                                                                               \\ 
\cline{1-1}\cline{3-5}
39                                                         &                                                          & SourceSearchCount     & Integer            & Number of search() functions HTML source code                                                                                               \\ 
\cline{1-1}\cline{3-5}
40                                                         &                                                          & ImageOnlyInForm       & Binary             & \begin{tabular}[c]{@{}l@{}}Check whether only image are present in HTML source code 
\end{tabular} \\
\hline
\end{tabular}
\end{table*}

\subsubsection{\textbf{\textit{Web-scrapped Features}}}
In this feature group, we would like to extract the web information of the URL's domain name and deal with the obfuscation problem. Some harmful URLs overlap lexical features with benign URLs. As a result, web scraping of URLs will offer us the required solution. URL-segmentation, deep-web features, host-based features, and content-based features are the four subgroups of our web-scrapped feature group. We want to identify ways to that spoof a domain name in \textbf{URL segmentation}. Whence, with Python, the domain name is sent as a query request in the Google search engine, and the first 60 results are collected for the process of hit counts. If the exact domain name exists, it is recorded as a count. This approach takes a little time, but it significantly influences the straightforward categorisation procedure. If similar name is found in netloc, it is an excellent chance to be a clean domain name. In \textbf{Deep-Web Features}, we create a list of 11 distinct types of typosquats of the searched domain name (see Table \ref{table:webscrappedfeaturegroup}) and examine all registered domain names among them. We also created a fuzzer for this step, which attempts to find typosquat domain names for each searched domain name. The phishing domains can be identified using above technique. We also computed the Lavenshtein distance and Shannon entropy with the most comparable query request results and saved it as a feature using the same method. The Lavenshtein distance between two words is the number of single-character modifications necessary to transform one term into another. In the \textbf{Host-based Features} subgroup, we included URL host-name attributes such as WHOIS information, IP address, location details, and country code. These include registration details, domain age, country code, location, and registrant company. As these attributes have identity-related qualities, they are recorded in numerical vectors with unique identities. For malicious URLs, the time-to-live from domain registration is nearly instantaneous. Many of them employed botnets for hosting for themselves on machines spread across various countries. As a result, host-based characteristics play an essential role in detecting malicious URLs. When the web page is completely downloaded, \textbf{Content-Based Features} are gained. Our research extracts HTML document-level information of the web page i.e. total images, the total links, and different tags from the source code. If malicious URLs remain undetected by URL-based features, then these attributes will function effectively in detecting threats by analysing the web page.

%
\subsection{\textbf{Empirical Study}}
\label{empiricalstudy}
This section describes our suggested supervised URL classification and unsupervised clustering approaches for detecting fraudulent advertisement URLs. We intend to assess if a given URL is benign or malicious by making binary classification problem. Consider the following collection of $N$ URLs: $(u_{1},y_{1}),...,(u_{n},y_{n})$. Here, $n=1,...N$ specifies a URL, and URL label is denoted by $y_{b}$, with $y=0$ denoting the benign URL and $y=1$ denoting the malicious URL. The first step is to represent the features $u_{n}\rightarrow a_{n}$, where $a_{n}$ is $n$-dimensional feature vector corresponding to the URL $u_{n}$. The problem can be expressed as a function $y_{n}=sign(L_{f}(a_{n}))$. We may reduce the overall number of errors in the classification process by minimising the loss function. We retrieved $89$ features from each URL and processed them for the classification phase based on the previous section. A balanced dataset of both legitimate and non-legitimate URLs is used to detect the fraudulent URL. We used $1000$ URLs from each of our six balanced merged datasets to train our four classifiers with appropriate parameters. Depending on the simulation task, we used a split size of around $0.7$ for training and $0.3$ for testing. These $89$ features will be used by classifiers to differentiate between clean and malicious URLs. In contrast to prior research that only considered a restricted element of the feature categories, we created a unique set of these highly influential feature categories and attempted to combine a good combination.
For supervised machine learning methods, we used Random Forest, Gradient Boost, AdaBoost, and XGBoost approaches. For each of our datasets, we examined the relative performance of all models. Then, to study the accuracy behaviour, we trained our models on mismatched datasets, training the complete one dataset and evaluating it on the other five. This procedure is carried out on all datasets with all model selections. We continued our research by testing our models on unlabeled datasets. We took all of the necessary criteria for the categorisation stage, as discussed in the previous section.

The process of grouping data points such similar data points are placed in same group and dissimilar from other data points in different groups is known as clustering. We employed the K-Means clustering method in this unsupervised learning strategy to better visualise malicious and benign data. The method will classify the objects into k groups of similarity, with an optimal value of k equal to $2$ expected for exact results. The euclidean distance will be used as a measurement to calculate that similarity. The Elbow approach is used to do this assignment. We cycle through the k values from $1$ to $9$ to calculate the distortion for each k value in the defined range. To get the optimal number of clusters, we must determine the value of k at the elbow or the point where the distortion begins to drop linearly.
%
\subsection{\textbf{\textit{ZOO - Adversarial Attack}}}
\label{zooadversarialattack}
This section will investigate our ensembles' vulnerability to adversarial attack. Unlike white-box attack approaches, which need the target network to be differentiable, in black-box ZOO attack the gradients are estimated uning zeroth order stochastic coordinate descent of the targeted network. Here, for this reason, we will select a loss function that is only dependent on the output of the targeted network in order to compute the gradient by using the finite difference method. We will use zeroth order optimization to solve the optimization problem. Here, for the input vector $a$, the loss function $L_{f}$ can be defined for output $O_{f}$ as
\begin{equation}
    L_{f}(a,b)=\max\{\max_{j\neq b}\log[O_{f}(a)]_{j}-\log[O_{f}(a)]_{b},-\rho\},
\end{equation}
where $\rho \geq0$ and $\log a \rightarrow -\infty$ whenever $a\rightarrow0$.
Because the $log$ function is a strictly monotonic function, $max=0$ follows. As a result, $a$ has the greatest confidence score for targeted class label $b$. Strictly monotonic functions are those function which for $x > y$ satisfy $f(x)>f(y)$. Here, the dominant influence is decreased by the $\log$ operator and maintain the confidence order due to monotonicity.
For untargeted attacks, if $a$ is classified as any different from original label $b_0$, then adversarial attack will be considered successful. For untargeted attack , the most probable pridicted class after eliminating $b$ is
\begin{equation}
    L_{f}(x)=\max\{\log[O_{f}(a)]_{b_0}-\max_{j\neq b_0}\log[O_{f}(a)]_{j},-\rho\},
\end{equation}
where original class label for $a$ is denoted by $b_0$, and the most probable predicted class is represented by $max_{j\neq b_0}\log[O_{f}(a)]_{j}$ other than $b_0$. The attack employs the optimization technique for the general $L_{f}$. If a function's symmetric derivative exists at point $a$, it is symmetrically differentiable at that point. Therefore, the symmetric difference quotient is:
\begin{equation}
    \hat{g_j}:=\frac{\partial L_{f}(a)}{\partial a_j}\approx \frac{L_{f}(a+ke_j)-L_{f}
(a-ke_j)}{2k},
\end{equation}
for gradient estimation. Here, we use $k$ as a very small constant and standard basis vector is denoted by $e_j$ where only the $j$th component have value $1$ and all other components equal to $0$. Although numerical precision is important, predicting the gradient precisely is typically not required for efficient adversarial attacks.
In the coordinate descent approach, one variable is chosen at random at each iteration and changed by minimising the objective function along that coordinate. We can obtain the estimated optimum delta by estimating the gradient and Hessian for $a_j$. Here, Hessian estimate is
\begin{equation}
    \hat{h_j}:= \frac{\partial ^2L_{f}(a)}{\partial ^2a^{2}_{jj}}\approx \frac{L_{f}(a+ke_j)-2L_{f}(a)+L_{f}(a-ke_j)}{k^2} ,
\end{equation}
The stochastic coordinate descent method in Algorithm \ref{alg:stochasticcoordinatedescent}  and the ADAM optimizer in Algorithm \ref{alg:zerothorderstochasticcoordinatedescentwithcoordinatewiseadam} increased the speed of the ZOO attack \cite{ref55}. The ZOO attack is more dependable to utilise than frequently used gradient-based techniques in this context due to its computational efficiency approximation and ease of implementation with only minor modifications.

\begin{algorithm}[h]
\caption{SCD - Stochastic Coordinate Descent}
\label{alg:stochasticcoordinatedescent}
\begin{algorithmic}
\STATE Initialise $l\leftarrow 0$;
\STATE {\textsc{\textbf{while}} {$l\leq K$} \textsc{\textbf{do}}}
\STATE \hspace{0.5cm}Choose $j(l)$ out of $\{1, ...,m \}$ with same probability
\STATE \hspace{0.5cm}Evaluate and update $\eta^{*}$
\STATE \hspace{2.6cm}${arg \min}_{\eta} L_{f}(a+\eta e_j)$
\STATE \hspace{0.5cm}Updating ${a_j}\leftarrow {a_j} +\eta^*$
\STATE \hspace{0.5cm}$l \leftarrow l+1$
\STATE {\textsc{\textbf{end while}}}
\end{algorithmic}
\end{algorithm}

\vspace{-10pt}
\begin{algorithm}[h]
\caption{Coordinate-wise ADAM Zeroth Order SCD}
\label{alg:zerothorderstochasticcoordinatedescentwithcoordinatewiseadam}
\begin{algorithmic}
\STATE 
\STATE {\textbf{Requirement:}} Set step size $h$, ADAM hyper-parameters $ \alpha_{1}=0.9$, ${\alpha_2=0.99}$, $\epsilon=10^{-8}$, and set ADAM states $N,\tau \in \mathbb{R}^{\wp}$
\STATE \hspace{1cm}Initialise $N\leftarrow 0$, $\tau \leftarrow 0$, $U\leftarrow 0$ 
\STATE {\textsc{\textbf{while}} {diverges} \textsc{\textbf{do}}}
\STATE \hspace{1cm}Choose $j(l)$ out of $\{1, ...,m \}$ with same probability
\STATE \hspace{1cm}Evaluate $U_j\leftarrow U_j+1$
\STATE \hspace{1cm}Approximate $g_j$ by using equation (3)
\STATE \hspace{1cm}Evaluate $N_j\leftarrow \alpha_1N_j+(1-\alpha_1)g_j$
\STATE \hspace{1cm}Evaluate $\tau_j \leftarrow \alpha_2 \tau_j+(1-\alpha_2)g_{j}^{2}$
\STATE \hspace{1cm}Evaluate $\hat{N_j}=N_j/(1-\alpha_{1}^{U_j}), \hat{\tau}=\tau_j/(1-\alpha_{2}^{U_j})$
\STATE
\STATE \hspace{2.6cm}$\eta^*=-h \frac{\hat{N_j}}{\sqrt{\hat{\tau_j}}+\epsilon}$
\STATE
\STATE \hspace{1cm}Update $a_j \leftarrow a_j+\eta^*$
\STATE {\textsc{\textbf{end while}}}
\end{algorithmic}
\end{algorithm}
\section{Results and Discussion}
\label{Results_and_Discussion}
Here, we will sum up and discuss our study results by inspecting the effectiveness of several classifiers, clustering techniques, and the effects of adversarial attacks on ensembles. To assess our suggested system, we run a set of rigorous tests on a desktop computer. The system's setup is Intel(R) Core(TM) Intel Core i5-4300M 2.6GHz, Ubuntu 20.04 LTS with 4GB RAM, and Python version 3.8.10.

\begin{table}[!ht]
\centering
\caption{Match Case Detection Result}
\label{table:matchcaseretectionresult}
\begin{tabular}{|c|c|c|c|c|c|} 
\hline
\multicolumn{6}{|c|}{{\cellcolor[rgb]{0.871,0.871,0.871}}\textbf{Random Forest}}                             \\ 
\hline
\textbf{Dataset}           & \textbf{Folds} & \textbf{Accuracy} & \textbf{Precision} & \textbf{FPR} & \textbf{FNR}  \\ 
\hline
\multirow{2}{*}{Dataset 1} & 5              & 98.72             & 98.98              & 0.0104       & 0.0152        \\ 
\cline{2-6}
                           & 10             & 99.13             & 99.42              & 0.0059       & 0.0115        \\ 
\hline
\multirow{2}{*}{Dataset 2} & 5              & 98.78             & 98.77              & 0.0120       & 0.0123        \\ 
\cline{2-6}
                           & 10             & 99.07             & 99.37              & 0.0062       & 0.0125        \\ 
\hline
\multirow{2}{*}{Dataset 3} & 5              & 99.06             & 99.37              & 0.0063       & 0.0125        \\ 
\cline{2-6}
                           & 10             & 99.13             & 98.86              & 0.0116       & 0.0057        \\ 
\hline
\multirow{2}{*}{Dataset 4} & 5              & 98.42             & 98.95              & 0.0107       & 0.0208        \\ 
\cline{2-6}
                           & 10             & 98.91             & 98.93              & 0.0112       & 0.0107        \\ 
\hline
\multirow{2}{*}{Dataset 5} & 5              & 98.84             & 98.86              & 0.0115       & 0.0114        \\ 
\cline{2-6}
                           & 10             & 99.18             & 98.92              & 0.0112       & 0.0057        \\ 
\hline
\multirow{2}{*}{Dataset 6} & 5              & 98.58             & 98.27              & 0.0175       & 0.0116        \\ 
\cline{2-6}
                           & 10             & 99.07             & 99.38              & 0.0061       & 0.0123        \\ 
\hline
\multicolumn{6}{|c|}{{\cellcolor[rgb]{0.871,0.871,0.871}}\textbf{AdaBoost}}                                  \\ 
\hline
\multirow{2}{*}{Dataset 1} & 5              & 98.98             & 98.97              & 0.0101       & 0.0103        \\ 
\cline{2-6}
                           & 10             & 99.23             & 98.97              & 0.0101       & 0.0052        \\ 
\hline
\multirow{2}{*}{Dataset 2} & 5              & 99.10             & 99.54              & 0.0048       & 0.0136        \\ 
\cline{2-6}
                           & 10             & 99.45             & 99.47              & 0.0052       & 0.0053        \\ 
\hline
\multirow{2}{*}{Dataset 3} & 5              & 99.13             & 98.58              & 0.0149       & 0.0048        \\ 
\cline{2-6}
                           & 10             & 99.56             & 99.53              & 0.0048       & 0.0047        \\ 
\hline
\multirow{2}{*}{Dataset 4} & 5              & 99.01             & 99.53              & 0.0050       & 0.0140        \\ 
\cline{2-6}
                           & 10             & 99.27             & 99.52              & 0.0050       & 0.0095        \\ 
\hline
\multirow{2}{*}{Dataset 5} & 5              & 99.04             & 99.53              & 0.0099       & 0.0094        \\ 
\cline{2-6}
                           & 10             & 99.35             & 99.07              & 0.0091       & 0.0047        \\ 
\hline
\multirow{2}{*}{Dataset 6} & 5              & 99.14             & 99.53              & 0.0047       & 0.0141        \\ 
\cline{2-6}
                           & 10             & 99.46             & 99.48              & 0.0051       & 0.0052        \\ 
\hline
\multicolumn{6}{|c|}{{\cellcolor[rgb]{0.871,0.871,0.871}}\textbf{Gradient Boost}}                                   \\ 
\hline
\multirow{2}{*}{Dataset 1} & 5              & 99.11             & 99.12              & 0.0091       & 0.0088        \\ 
\cline{2-6}
                           & 10             & 99.31             & 99.55              & 0.0045       & 0.0090        \\ 
\hline
\multirow{2}{*}{Dataset 2} & 5              & 99.14             & 99.56              & 0.0044       & 0.0131        \\ 
\cline{2-6}
                           & 10             & 99.46             & 99.38              & 0.0041       & 0.0085        \\ 
\hline
\multirow{2}{*}{Dataset 3} & 5              & 99.19             & 99.57              & 0.0042       & 0.0127        \\ 
\cline{2-6}
                           & 10             & 99.60             & 99.60              & 0.0041       & 0.0081        \\ 
\hline
\multirow{2}{*}{Dataset 4} & 5              & 99.04             & 99.14              & 0.0062       & 0.0124        \\ 
\cline{2-6}
                           & 10             & 99.39             & 99.59              & 0.0043       & 0.0043        \\ 
\hline
\multirow{2}{*}{Dataset 5} & 5              & 99.11             & 99.55              & 0.0045       & 0.0131        \\ 
\cline{2-6}
                           & 10             & 99.38             & 99.15              & 0.0083       & 0.0043        \\ 
\hline
\multirow{2}{*}{Dataset 6} & 5              & 99.29             & 99.53              & 0.0047       & 0.0093        \\ 
\cline{2-6}
                           & 10             & 99.54             & 99.55              & 0.0047       & 0.0045        \\ 
\hline
\multicolumn{6}{|c|}{{\cellcolor[rgb]{0.871,0.871,0.871}}\textbf{XGBoost}}                                   \\ 
\hline
\multirow{2}{*}{Dataset 1} & 5              & 99.26             & 99.25              & 0.0073       & 0.0075        \\ 
\cline{2-6}
                           & 10             & 99.59             & 99.59              & 0.0041       & 0.0041        \\ 
\hline
\multirow{2}{*}{Dataset 2} & 5              & 99.24             & 99.25              & 0.0077       & 0.0075        \\ 
\cline{2-6}
                           & 10             & 99.47             & 99.63              & 0.0036       & 0.0073        \\ 
\hline
\multirow{2}{*}{Dataset 3} & 5              & 99.27             & 99.25              & 0.0071       & 0.0075        \\ 
\cline{2-6}
                           & 10             & 99.63             & 99.63              & 0.0037       & 0.0037        \\ 
\hline
\multirow{2}{*}{Dataset 4} & 5              & 99.15             & 99.14              & 0.0085       & 0.0086        \\ 
\cline{2-6}
                           & 10             & 99.58             & 99.58              & 0.0041       & 0.0042        \\ 
\hline
\multirow{2}{*}{Dataset 5} & 5              & 99.35             & 99.15              & 0.0089       & 0.0043        \\ 
\cline{2-6}
                           & 10             & 99.42             & 99.24              & 0.0078       & 0.0038        \\ 
\hline
\multirow{2}{*}{Dataset 6} & 5              & 99.45             & 99.25              & 0.0076       & 0.0038        \\ 
\cline{2-6}
                           & 10             & 99.63             & 99.63              & 0.0037       & 0.0037        \\
\hline
\end{tabular}
\end{table}

\subsection{\textbf{\textit{Experimental Results}}}
We pick 1000 URLs from each balanced dataset with an equal number of malicious and benign advertisement URLs for the experiment. The experiment is performed on each of our four classifiers with 5 and 10 folds. We also consider the runtime complexity, which grows as the number of folds increases. It is a problem throughout the training stage but not during testing. Each of the six datasets took roughly 150 minutes to process. However, the model's remarkable accuracy is worth it in the end. We used Python's Sciket-learn package and the GridSearchCV library function to cycle over predefined hyperparameters, selecting 1, 100, 200, 500, 1000, and 1500 \textit{n\_estimators} to fit on our training set. We got more precise results as the number of trees increased. The sampled datasets were examined using the Random Forest model, and we noticed a gain in accuracy and precision as we moved to AdaBoost, Gradient Boost, and XGBoost, as well as a drop in FPR and FNR. We discovered that the ten folds of the XGBoost classifier had the best detection accuracy. We created the evaluation matrix using the following formulas.\\
\textit{FPR:} The false positive rate is calculated as the proportion of malicious data that is mistakenly recognised as benign.\\
\textit{FNR:} The false negative rate is the proportion of data that is wrongly classified as non-malicious but is, in fact, malicious.

\begin{table*}

\centering
\caption{Mismatch Case Detection Result}
\label{table:mismatchcasedetectionresult}
\begin{tabular}{|c|c|c|c|c|c|c|c|c|c|c|c|} 
\hline
\rowcolor[rgb]{0.871,0.871,0.871} \multicolumn{6}{|c|}{\textbf{Random Forest}}                                                                                             & \multicolumn{6}{c|}{\textbf{AdaBoost}}                                                                                                                                              \\ 
\hline
\textbf{Trained}           & \textbf{Tested}                            & \textbf{Accuracy}          & \textbf{Precision}          & \textbf{FPR}          & \textbf{FNR}          & \textbf{\textbf{Trained}}  & \textbf{\textbf{Tested}}                   & \textbf{\textbf{Accuracy}} & \textbf{\textbf{Precision}} & \textbf{\textbf{FPR}} & \textbf{\textbf{FNR}}  \\ 
\hline
\multirow{5}{*}{Dataset 1} & Dataset 2                                  & 98.69                      & 98.77                       & 0.0121                & 0.0143                & \multirow{5}{*}{Dataset 1} & Dataset 2                                  & 98.70                      & 98.89                       & 0.0131                & 0.0129                 \\ 
\cline{2-6}\cline{8-12}
                           & Dataset 3                                  & 98.89                      & 99.18                       & 0.0083                & 0.0143                &                            & Dataset 3                                  & 98.93                      & 98.81                       & 0.0140                & 0.0079                 \\ 
\cline{2-6}\cline{8-12}
                           & Dataset 4                                  & 98.81                      & 99.09                       & 0.0091                & 0.0150                &                            & Dataset 4                                  & 98.84                      & 98.81                       & 0.0140                & 0.0096                 \\ 
\cline{2-6}\cline{8-12}
                           & Dataset 5                                  & 98.92                      & 98.69                       & 0.0129                & 0.0093                &                            & Dataset 5                                  & 98.99                      & 98.83                       & 0.0142                & 0.0068                 \\ 
\cline{2-6}\cline{8-12}
                           & Dataset 6                                  & 98.65                      & 98.79                       & 0.0119                & 0.0154                &                            & Dataset 6                                  & 98.77                      & 98.89                       & 0.0131                & 0.0117                 \\ 
\hline
\multirow{5}{*}{Dataset 2} & Dataset 1                                  & 98.78                      & 98.95                       & 0.0103                & 0.0141                & \multirow{5}{*}{Dataset 2} & Dataset 1                                  & 98.86                      & 99.06                       & 0.0111                & 0.0117                 \\ 
\cline{2-6}\cline{8-12}
                           & Dataset 3                                  & 98.70                      & 98.88                       & 0.0111                & 0.0150                &                            & Dataset 3                                  & 98.92                      & 99.12                       & 0.0103                & 0.0105                 \\ 
\cline{2-6}\cline{8-12}
                           & Dataset 4                                  & 98.70                      & 98.75                       & 0.0123                & 0.0137                &                            & Dataset 4                                  & 98.96                      & 99.29                       & 0.0084                & 0.0122                 \\ 
\cline{2-6}\cline{8-12}
                           & Dataset 5                                  & 98.56                      & 98.16                       & 0.0181                & 0.0106                &                            & Dataset 5                                  & 98.87                      & 99.25                       & 0.0088                & 0.0135                 \\ 
\cline{2-6}\cline{8-12}
                           & Dataset 6                                  & 98.68                      & 98.41                       & 0.0157                & 0.0104                &                            & Dataset 6                                  & 98.76                      & 98.92                       & 0.0127                & 0.0121                 \\ 
\hline
\multirow{5}{*}{Dataset 3} & Dataset 1                                  & 98.83                      & 98.59                       & 0.0138                & 0.0094                & \multirow{5}{*}{Dataset 3} & Dataset 1                                  & 98.88                      & 99.20                       & 0.0082                & 0.0141                 \\ 
\cline{2-6}\cline{8-12}
                           & Dataset 2                                  & 98.66                      & 98.81                       & 0.0118                & 0.0147                &                            & Dataset 2                                  & 98.68                      & 99.00                       & 0.0103                & 0.0161                 \\ 
\cline{2-6}\cline{8-12}
                           & Dataset 4                                  & 98.39                      & 98.04                       & 0.0194                & 0.0128                &                            & Dataset 4                                  & 98.59                      & 98.99                       & 0.0103                & 0.0181                 \\ 
\cline{2-6}\cline{8-12}
                           & Dataset 5                                  & 98.61                      & 98.37                       & 0.0161                & 0.0117                &                            & Dataset 5                                  & 98.70                      & 99.19                       & 0.0082                & 0.0177                 \\ 
\cline{2-6}\cline{8-12}
                           & Dataset 6                                  & 98.77                      & 98.59                       & 0.0140                & 0.0106                &                            & Dataset 6                                  & 98.95                      & 98.71                       & 0.0131                & 0.0081                 \\ 
\hline
\multirow{5}{*}{Dataset 4} & Dataset 1                                  & 98.53                      & 98.64                       & 0.0166                & 0.0131                & \multirow{5}{*}{Dataset 4} & Dataset 1                                  & 98.64                      & 99.11                       & 0.0090                & 0.018                  \\ 
\cline{2-6}\cline{8-12}
                           & Dataset 2                                  & 98.43                      & 98.58                       & 0.0174                & 0.0143                &                            & Dataset 2                                  & 98.59                      & 99.31                       & 0.0070                & 0.0211                 \\ 
\cline{2-6}\cline{8-12}
                           & Dataset 3                                  & 98.59                      & 98.58                       & 0.0139                & 0.0143                &                            & Dataset 3                                  & 98.89                      & 99.14                       & 0.0085                & 0.0141                 \\ 
\cline{2-6}\cline{8-12}
                           & Dataset 5                                  & 98.71                      & 98.98                       & 0.0099                & 0.0160                &                            & Dataset 5                                  & 98.80                      & 99.37                       & 0.0064                & 0.0174                 \\ 
\cline{2-6}\cline{8-12}
                           & Dataset 6                                  & 99.04                      & 98.84                       & 0.0113                & 0.0079                &                            & Dataset 6                                  & 98.04                      & 99.23                       & 0.0078                & 0.0114                 \\ 
\hline
\multirow{5}{*}{Dataset 5} & Dataset 1                                  & 98.67                      & 98.60                       & 0.0137                & 0.0128                & \multirow{5}{*}{Dataset 5} & Dataset 1                                  & 98.70                      & 98.44                       & 0.0158                & 0.0103                 \\ 
\cline{2-6}\cline{8-12}
                           & Dataset 2                                  & 98.70                      & 98.56                       & 0.0142                & 0.0118                &                            & Dataset 2                                  & 98.90                      & 98.64                       & 0.0135                & 0.0087                 \\ 
\cline{2-6}\cline{8-12}
                           & Dataset 3                                  & 98.49                      & 98.13                       & 0.0184                & 0.0118                &                            & Dataset 3                                  & 98.66                      & 98.39                       & 0.0164                & 0.0101                 \\ 
\cline{2-6}\cline{8-12}
                           & Dataset 4                                  & 98.63                      & 98.27                       & 0.0170                & 0.0104                &                            & Dataset 4                                  & 98.78                      & 98.59                       & 0.0143                & 0.0113                 \\ 
\cline{2-6}\cline{8-12}
                           & Dataset 6                                  & 98.73                      & 98.33                       & 0.0164                & 0.0090                &                            & Dataset 6                                  & 98.79                      & 98.79                       & 0.0123                & 0.0118                 \\ 
\hline
\multirow{5}{*}{Dataset 6} & Dataset 1                                  & 98.38                      & 98.05                       & 0.0229                & 0.0105                & \multirow{5}{*}{Dataset 6} & Dataset 1                                  & 98.52                      & 97.99                       & 0.0293                & 0.0094                 \\ 
\cline{2-6}\cline{8-12}
                           & Dataset 2                                  & 98.34                      & 98.22                       & 0.0209                & 0.0129                &                            & Dataset 2                                  & 98.71                      & 98.20                       & 0.0182                & 0.0075                 \\ 
\cline{2-6}\cline{8-12}
                           & Dataset 3                                  & 97.80                      & 98.05                       & 0.0228                & 0.0212                &                            & Dataset 3                                  & 98.45                      & 98.12                       & 0.0191                & 0.0119                 \\ 
\cline{2-6}\cline{8-12}
                           & Dataset 4                                  & 98.07                      & 98.55                       & 0.0170                & 0.0212                &                            & Dataset 4                                  & 98.68                      & 98.19                       & 0.0184                & 0.0081                 \\ 
\cline{2-6}\cline{8-12}
                           & Dataset 5                                  & 98.16                      & 98.39                       & 0.0190                & 0.0179                &                            & Dataset 5                                  & 98.58                      & 98.39                       & 0.0164                & 0.0121                 \\ 
\hline
\rowcolor[rgb]{0.871,0.871,0.871} \multicolumn{6}{|c|}{\textbf{Gradient Boost}}                                                                                                    & \multicolumn{6}{c|}{\textbf{XGBoost}}                                                                                                                                               \\ 
\hline
\textbf{\textbf{Trained}}  & \textbf{\textbf{\textbf{\textbf{Tested}}}} & \textbf{\textbf{Accuracy}} & \textbf{\textbf{Precision}} & \textbf{\textbf{FPR}} & \textbf{\textbf{FNR}} & \textbf{\textbf{Trained}}  & \textbf{\textbf{\textbf{\textbf{Tested}}}} & \textbf{\textbf{Accuracy}} & \textbf{\textbf{Precision}} & \textbf{\textbf{FPR}} & \textbf{\textbf{FNR}}  \\ 
\hline
\multirow{5}{*}{Dataset 1} & Dataset 2                                  & 98.58                      & 98.84                       & 0.0118                & 0.0129                & \multirow{5}{*}{Dataset 1} & Dataset 2                                  & 98.91                      & 98.81                       & 0.0116                & 0.0102                 \\ 
\cline{2-6}\cline{8-12}
                           & Dataset 3                                  & 98.96                      & 98.75                       & 0.0127                & 0.0081                &                            & Dataset 3                                  & 99.10                      & 99.14                       & 0.0084                & 0.0096                 \\ 
\cline{2-6}\cline{8-12}
                           & Dataset 4                                  & 98.95                      & 98.80                       & 0.0122                & 0.0089                &                            & Dataset 4                                  & 99.07                      & 98.85                       & 0.0112                & 0.0073                 \\ 
\cline{2-6}\cline{8-12}
                           & Dataset 5                                  & 99.05                      & 99.22                       & 0.0081                & 0.0109                &                            & Dataset 5                                  & 99.13                      & 98.93                       & 0.0104                & 0.0069                 \\ 
\cline{2-6}\cline{8-12}
                           & Dataset 6                                  & 98.87                      & 98.97                       & 0.0100                & 0.0123                &                            & Dataset 6                                  & 99.01                      & 98.97                       & 0.0100                & 0.0097                 \\ 
\hline
\multirow{5}{*}{Dataset 2} & Dataset 1                                  & 98.95                      & 98.89                       & 0.0109                & 0.0101                & \multirow{5}{*}{Dataset 2} & Dataset 1                                  & 99.02                      & 99.29                       & 0.0070                & 0.0127                 \\ 
\cline{2-6}\cline{8-12}
                           & Dataset 3                                  & 99.11                      & 98.87                       & 0.0100                & 0.0080                &                            & Dataset 3                                  & 99.15                      & 99.26                       & 0.0072                & 0.0096                 \\ 
\cline{2-6}\cline{8-12}
                           & Dataset 4                                  & 99.12                      & 98.98                       & 0.0099                & 0.0076                &                            & Dataset 4                                  & 99.16                      & 99.05                       & 0.0092                & 0.0076                 \\ 
\cline{2-6}\cline{8-12}
                           & Dataset 5                                  & 98.12                      & 98.12                       & 0.0102                & 0.0117                &                            & Dataset 5                                  & 99.06                      & 98.86                       & 0.0092                & 0.0076                 \\ 
\cline{2-6}\cline{8-12}
                           & Dataset 6                                  & 98.91                      & 98.57                       & 0.0139                & 0.0076                &                            & Dataset 6                                  & 99.03                      & 99.01                       & 0.0096                & 0.0097                 \\ 
\hline
\multirow{5}{*}{Dataset 3} & Dataset 1                                  & 98.97                      & 98.59                       & 0.0138                & 0.0066                & \multirow{5}{*}{Dataset 3} & Dataset 1                                  & 99.08                      & 98.98                       & 0.0100                & 0.0083                 \\ 
\cline{2-6}\cline{8-12}
                           & Dataset 2                                  & 98.84                      & 98.44                       & 0.0151                & 0.0079                &                            & Dataset 2                                  & 98.98                      & 99.08                       & 0.0090                & 0.0114                 \\ 
\cline{2-6}\cline{8-12}
                           & Dataset 4                                  & 98.87                      & 98.57                       & 0.0140                & 0.0086                &                            & Dataset 4                                  & 98.95                      & 99.03                       & 0.0095                & 0.0115                 \\ 
\cline{2-6}\cline{8-12}
                           & Dataset 5                                  & 98.94                      & 98.85                       & 0.0113                & 0.0100                &                            & Dataset 5                                  & 99.02                      & 99.15                       & 0.0083                & 0.0114                 \\ 
\cline{2-6}\cline{8-12}
                           & Dataset 6                                  & 99.15                      & 98.74                       & 0.0123                & 0.0046                &                            & Dataset 6                                  & 99.21                      & 99.10                       & 0.0088                & 0.0069                 \\ 
\hline
\multirow{5}{*}{Dataset 4} & Dataset 1                                  & 98.75                      & 98.54                       & 0.0143                & 0.0106                & \multirow{5}{*}{Dataset 4} & Dataset 1                                  & 98.92                      & 98.74                       & 0.0123                & 0.0074                 \\ 
\cline{2-6}\cline{8-12}
                           & Dataset 2                                  & 98.79                      & 98.57                       & 0.0140                & 0.0103                &                            & Dataset 2                                  & 98.85                      & 98.65                       & 0.0132                & 0.0097                 \\ 
\cline{2-6}\cline{8-12}
                           & Dataset 3                                  & 98.91                      & 98.44                       & 0.0152                & 0.0065                &                            & Dataset 3                                  & 99.01                      & 98.65                       & 0.0131                & 0.0065                 \\ 
\cline{2-6}\cline{8-12}
                           & Dataset 5                                  & 98.97                      & 98.79                       & 0.0118                & 0.0087                &                            & Dataset 5                                  & 99.10                      & 98.94                       & 0.0103                & 0.0077                 \\ 
\cline{2-6}\cline{8-12}
                           & Dataset 6                                  & 99.16                      & 98.89                       & 0.0109                & 0.0058                &                            & Dataset 6                                  & 99.20                      & 99.38                       & 0.0060                & 0.0102                 \\ 
\hline
\multirow{5}{*}{Dataset 5} & Dataset 1                                  & 98.89                      & 99.15                       & 0.0083                & 0.0139                & \multirow{5}{*}{Dataset 5} & Dataset 1                                  & 99.05                      & 99.03                       & 0.0095                & 0.0095                 \\ 
\cline{2-6}\cline{8-12}
                           & Dataset 2                                  & 98.99                      & 98.75                       & 0.0122                & 0.0078                &                            & Dataset 2                                  & 99.14                      & 99.04                       & 0.0093                & 0.0078                 \\ 
\cline{2-6}\cline{8-12}
                           & Dataset 3                                  & 99.04                      & 99.15                       & 0.0083                & 0.0110                &                            & Dataset 3                                  & 99.15                      & 98.74                       & 0.0123                & 0.0045                 \\ 
\cline{2-6}\cline{8-12}
                           & Dataset 4                                  & 98.95                      & 98.94                       & 0.0104                & 0.0107                &                            & Dataset 4                                  & 99.07                      & 99.01                       & 0.0097                & 0.0090                 \\ 
\cline{2-6}\cline{8-12}
                           & Dataset 6                                  & 98.96                      & 98.95                       & 0.0103                & 0.0106                &                            & Dataset 6                                  & 99.03                      & 98.82                       & 0.0116                & 0.0078                 \\ 
\hline

\multirow{5}{*}{Dataset 6} & Dataset 1                                  & 98.75                      & 98.26                       & 0.0170                & 0.0078                & \multirow{5}{*}{Dataset 6} & Dataset 1                                  & 98.85                      & 98.62                       & 0.0134                & 0.0095                 \\ 
\cline{2-6}\cline{8-12}
                           & Dataset 2                                  & 98.85                      & 98.66                       & 0.0131                & 0.0099                &                            & Dataset 2                                  & 98.93                      & 98.77                       & 0.0120                & 0.0094                 \\ 
\cline{2-6}\cline{8-12}
                           & Dataset 3                                  & 98.64                      & 98.44                       & 0.0152                & 0.0121                &                            & Dataset 3                                  & 98.82                      & 98.94                       & 0.0103                & 0.0134                 \\ 
\cline{2-6}\cline{8-12}
                           & Dataset 4                                  & 98.92                      & 98.46                       & 0.0150                & 0.0065                &                            & Dataset 4                                  & 99.05                      & 98.74                       & 0.0123                & 0.0065                 \\ 
\cline{2-6}\cline{8-12}
                           & Dataset 5                                  & 98.77                      & 98.36                       & 0.0160                & 0.0086                &                            & Dataset 5                                  & 98.90                      & 98.88                       & 0.0109                & 0.0110                 \\
\hline

\end{tabular}
\end{table*}

\begin{equation*}
     FPR=\frac{FP}{FP+TN}, \hspace{0.5cm} FNR=\frac{FN}{TP+FN}
\end{equation*}
\textit{Accuracy}: Accuracy is calculated as the ratio of accurately predicted examples to total examples.\vspace{3pt}\\
\textit{Precision}: The percentage of malicious data that was successfully categorised out of all malicious URLs labelled by the classifier.
\begin{footnotesize}
\begin{equation*}
    Precision \&=\frac{TP}{TP+FP},\hspace{0.2cm}
    Accuracy \&=\frac{TP+TN}{TP+TN+FP+FN}
\end{equation*}
\end{footnotesize}

\subsubsection{\bf Detection Results}We have discussed the accuracy and precision of our ensembles by using the confusion matrix in this section with the discussed machine learning techniques. 

\begin{itemize}
\item{\textbf{Matched Case}}\\
In the matched scenario, the classifier is trained and tested on a single dataset. The results for all classifiers are shown in Table \ref{table:matchcaseretectionresult}.

\item{\textbf{Mismatched Case}}\\
The classifier is trained on one full dataset and evaluated on the remaining five datasets in the mismatched case. Table \ref{table:mismatchcasedetectionresult} shows the results tables for various classifiers for mismatched datasets.

\end{itemize}

\subsubsection{\bf Clustering Results}
In this section of our research, we choose cluster formation to distinguish between two types of malicious and non-malicious URLs. We employ the K-Means clustering technique for visual comprehension for this purpose. We determine the number of clusters necessary to cluster the dataset using the elbow technique of K-Means. We discover a gap between the two groups and make a clear distinction. This creation is seen in Fig. \ref{fig:clusteringresults}, where red represents benign and blue represents malicious.



\begin{figure}[H]
\centering
\includegraphics[width=3.5in]{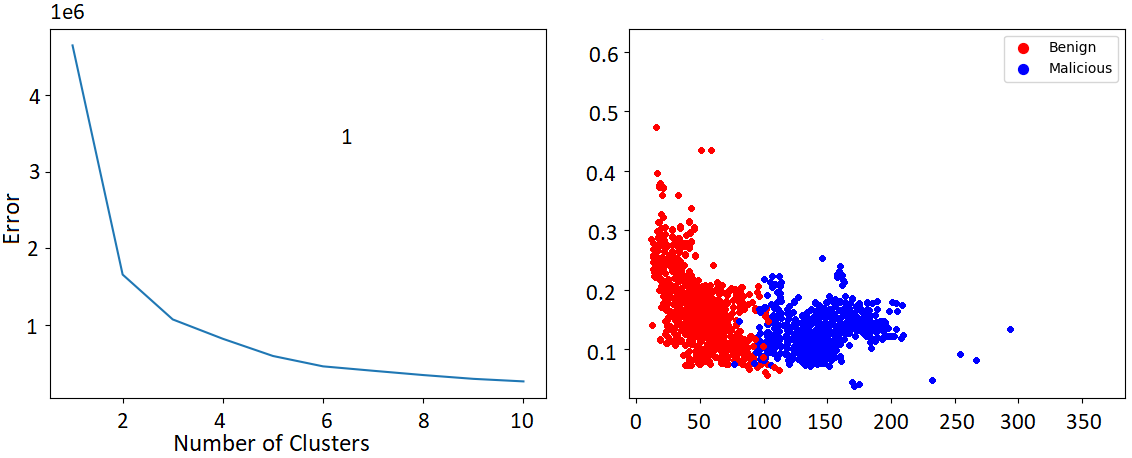}
\caption{Elbow method and K-Means result.}
\label{fig:elbowmethodandkmeansresult}
\end{figure}

\subsubsection{\bf Adversarial Attack Results}
Here, we conduct technical investigations on our model and investigate the impact of adversarial training on our models. We used the ZOO attack to evaluate the susceptibility of various models. We trained our model first and then used the ZOO attack in the testing phase. The result for the ZOO attack is shown in Table \ref{table:zooattackresult}.

\vspace{-5pt}
\begin{table}[h]
\centering
\caption{ZOO Attack Result}
\label{table:zooattackresult}
\begin{tabular}{|@{}c|@{}c@{}|c|c|c|c|} 
\hline
\multirow{2}{*}{\textbf{Dataset}} & \multirow{2}{*}{\textbf{Confidence}} & \multicolumn{4}{c|}{\textbf{Attack Accuracy Rate}}                                                                                                                                           \\ 
\cline{3-6}
                                  &                                      & \begin{tabular}[c]{@{}c@{}}\textbf{Random}\\\textbf{Forest}\end{tabular} & \textbf{AdaBoost} & \begin{tabular}[c]{@{}c@{}}\textbf{Gradient}\\\textbf{Boost}\end{tabular} & \textbf{XGBoost}  \\ 
\hline \hline
\multirow{2}{*}{Dataset 1}        & 50                                   & 95.85                                                                    & 95.62             & 95.30                                                                     & 93.15             \\ 
\cline{2-6}
                                  & 100                                  & 94.33                                                                    & 95.31             & 94.87                                                                     & 93.86             \\ 
\hline
\multirow{2}{*}{Dataset 2}        & 50                                   & 95.18                                                                    & 94.50             & 94.41                                                                     & 95.16             \\ 
\cline{2-6}
                                  & 100                                  & 93.80                                                                    & 94.33             & 94.66                                                                     & 93.05             \\ 
\hline
\multirow{2}{*}{Dataset 3}        & 50                                   & 96.05                                                                    & 96.18             & 95.30                                                                     & 94.82             \\ 
\cline{2-6}
                                  & 100                                  & 96.00                                                                    & 95.95             & 95.73                                                                     & 93.78             \\ 
\hline
\multirow{2}{*}{Dataset 4}        & 50                                   & 96.67                                                                    & 95.31             & 94.44                                                                     & 93.70             \\ 
\cline{2-6}
                                  & 100                                  & 96.50                                                                    & 95.12             & 94.66                                                                     & 93.40             \\ 
\hline
\multirow{2}{*}{Dataset 5}        & 50                                   & 95.74                                                                    & 95.29             & 94.30                                                                     & 93.64             \\ 
\cline{2-6}
                                  & 100                                  & 94.60                                                                    & 94.56             & 94.30                                                                     & 92.61             \\ 
\hline
\multirow{2}{*}{Dataset 6}        & 50                                   & 95.20                                                                    & 95.12             & 95.35                                                                     & 92.20             \\ 
\cline{2-6}
                                  & 100                                  & 95.05                                                                    & 94.66             & 94.88                                                                     & 92.18             \\
\hline
\end{tabular}
\end{table}

\subsection{\textbf{Discussion}}
The combination of lexical and web-scrapped features we collected is a good foundation for our analysis. Various machine learning techniques used for categorization give us light and profound output. We found that the XGBoost classifier with a maximum depth of ten yields the best outcomes in the classification process. When compared to other classifiers, XGBoost has the highest accuracy and precision, followed by Gradient Boost, AdaBoost, and Random Forest. If the number of folds in the simulation work increases, the FPR and FNR drop. The model grows more sophisticated and overfits the data as the maximum depth of the tree is increased. The same thing occurs with the other classifiers. Our findings and recommended techniques improve the accuracy of prediction and detection. The accuracy we discovered is substantially greater than prior efforts, and deploying machine learning systems in the identification of rogue advertisement URLs is a good concept. The visual comprehension in the elbow method cluster formation recommends the appropriate categories of data and low overlaps of both classes. The results of the adversarial attack indicate a comparison of the ensemble's susceptibility. We discovered that better predictive models are less vulnerable to adversarial examples. Due to a lack of attack tools to test their resilience, studying robustness training techniques for tree ensemble models has been challenging. Our method can serve as a benchmark tool for robustness evaluation, stimulating research in tree ensemble robustness. The research on lowering time complexity and developing appropriate defence strategies will benefit our studies.
\section{Conclusion and Future Work}
\label{CONCLUSION_AND_FUTURE WORK}
In this paper, we developed a framework for identifying fraudulent advertisement URLs using innovative feature extraction, and we created a detection system with the highest accuracy results. We also looked into the vulnerability of various ensembles for adversarial training. We investigated novel feature extraction using an unconventional combination of six feature classes, including various advanced features. Most of the existing approaches were based on only the traditional features set like a bag of words, which were insufficient to make the detection system sufficiently efficient. We focused on handling the unseen features to test the URL's maliciousness. Our classifiers optimised the accuracy, which we can see in the clustering visualisation. The machine learning approaches we used are ideal for the purpose. 

Although the web-scrapped features extraction process was a little time-consuming due to a number of ample features, we would take it to minimise with more resources and tactics in the upcoming time. We will also deploy the state of the art adversarial examples and compare the vulnerability and defence approaches in our future work.


\bibliographystyle{IEEEtran}
\bibliography{References.bib}

\begin{thebibliography}{10}
\providecommand{\url}[1]{#1}
\csname url@samestyle\endcsname
\providecommand{\newblock}{\relax}
\providecommand{\bibinfo}[2]{#2}
\providecommand{\BIBentrySTDinterwordspacing}{\spaceskip=0pt\relax}
\providecommand{\BIBentryALTinterwordstretchfactor}{4}
\providecommand{\BIBentryALTinterwordspacing}{\spaceskip=\fontdimen2\font plus
\BIBentryALTinterwordstretchfactor\fontdimen3\font minus
  \fontdimen4\font\relax}
\providecommand{\BIBforeignlanguage}[2]{{%
\expandafter\ifx\csname l@#1\endcsname\relax
\typeout{** WARNING: IEEEtran.bst: No hyphenation pattern has been}%
\typeout{** loaded for the language `#1'. Using the pattern for}%
\typeout{** the default language instead.}%
\else
\language=\csname l@#1\endcsname
\fi
#2}}
\providecommand{\BIBdecl}{\relax}
\BIBdecl

\bibitem{ref3}
J.~Hong, ``The state of phishing attacks,'' \emph{Communications of the ACM},
  vol.~55, no.~1, pp. 74--81, 2012.

\bibitem{ref1}
D.~Sahoo, C.~Liu, and S.~C. Hoi, ``Malicious url detection using machine
  learning: A survey,'' \emph{arXiv preprint arXiv:1701.07179}, 2017.

\bibitem{ref4}
A.~Blum, B.~Wardman, T.~Solorio, and G.~Warner, ``Lexical feature based
  phishing url detection using online learning,'' in \emph{Proceedings of the
  3rd ACM Workshop on Artificial Intelligence and Security}, 2010, pp. 54--60.

\bibitem{ref49}
``{GitHub Code},''
  \url{https://github.com/ehsannowroozi/Sec_Classifying_URL_Detection},
  accessed: 2022-04-07.

\bibitem{ref47}
``{VirusTotal},'' \url{https://www.virustotal.com/gui/home/url}, accessed:
  2022-03-24.

\bibitem{ref38}
``{Alexa Dataset},'' \url{https://www.kaggle.com/datasets/cheedcheed/top1m},
  accessed: 2021-08-24.

\bibitem{ref39}
``{Crowdflower Dataset},''
  \url{https://data.world/crowdflower/url-categorization}, accessed:
  2022-03-24.

\bibitem{ref40}
``{DMOZ Dataset},''
  \url{https://www.kaggle.com/datasets/shawon10/url-classification-dataset-dmoz},
  accessed: 2022-03-24.

\bibitem{ref41}
``{Benign and Malicious Set URL},''
  \url{https://www.kaggle.com/datasets/siddharthkumar25/malicious-and-benign-urls},
  accessed: 2022-03-24.

\bibitem{ref42}
``{Benign and Malicious Data URL Dataset},''
  \url{https://www.kaggle.com/antonyj453/urldataset#data.csv}, accessed:
  2022-03-24.

\bibitem{ref43}
``{ISCX-URL-2016 Dataset},''
  \url{https://www.unb.ca/cic/datasets/url-2016.html}, accessed: 2022-03-24.

\bibitem{ref36}
M.~S.~I. Mamun, M.~A. Rathore, A.~H. Lashkari, N.~Stakhanova, and A.~A.
  Ghorbani, ``Detecting malicious urls using lexical analysis,'' in
  \emph{International Conference on Network and System Security}.\hskip 1em
  plus 0.5em minus 0.4em\relax Springer, 2016, pp. 467--482.

\bibitem{ref45}
``{Phishtank Dataset},'' \url{https://www.phishtank.com/developer_info.php},
  accessed: 2022-03-24.

\bibitem{ref46}
``{Phishstrom Dataset},''
  \url{https://research.aalto.fi/en/datasets/phishstorm-phishing-legitimate-url-dataset},
  accessed: 2022-03-24.

\bibitem{ref37}
S.~Marchal, J.~Fran{\c{c}}ois, R.~State, and T.~Engel, ``Phishstorm: Detecting
  phishing with streaming analytics,'' \emph{IEEE Transactions on Network and
  Service Management}, vol.~11, no.~4, pp. 458--471, 2014.

\bibitem{ref44}
``{Phishing Site URL Dataset},''
  \url{https://www.kaggle.com/datasets/taruntiwarihp/phishing-site-urls},
  accessed: 2022-03-24.

\bibitem{IEEEtnsm2022}
E.~Nowroozi, Y.~Mekdad, M.~H. Berenjestanaki, M.~Conti, and A.~EL~Fergougui,
  ``Demystifying the transferability of adversarial attacks in computer
  networks,'' \emph{IEEE Transactions on Network and Service Management}, pp.
  1--1, 2022.

\bibitem{ref53}
C.~Szegedy, W.~Zaremba, I.~Sutskever, J.~Bruna, D.~Erhan, I.~Goodfellow, and
  R.~Fergus, ``Intriguing properties of neural networks,'' \emph{arXiv preprint
  arXiv:1312.6199}, 2013.

\bibitem{ref48}
I.~J. Goodfellow, J.~Shlens, and C.~Szegedy, ``Explaining and harnessing
  adversarial examples,'' \emph{arXiv preprint arXiv:1412.6572}, 2014.

\bibitem{ref54}
E.~Nowroozi, A.~Dehghantanha, R.~M. Parizi, and K.-K.~R. Choo, ``A survey of
  machine learning techniques in adversarial image forensics,'' \emph{Computers
  \& Security}, vol. 100, p. 102092, 2021.

\bibitem{nowroozi2020machine}
E.~Nowroozi, M.~Barni, and B.~Tondi, ``Machine learning techniques for image
  forensics in adversarial setting,'' Ph.D. dissertation, 2020.

\bibitem{ref21}
N.~Carlini and D.~Wagner, ``Audio adversarial examples: Targeted attacks on
  speech-to-text,'' in \emph{IEEE Security and Privacy Workshops (SPW)}.\hskip
  1em plus 0.5em minus 0.4em\relax IEEE, 2018, pp. 1--7.

\bibitem{ref10}
P.~Prakash, M.~Kumar, R.~R. Kompella, and M.~Gupta, ``Phishnet: predictive
  blacklisting to detect phishing attacks,'' in \emph{2010 Proceedings IEEE
  INFOCOM}.\hskip 1em plus 0.5em minus 0.4em\relax IEEE, 2010, pp. 1--5.

\bibitem{ref9}
Y.~He, Z.~Zhong, S.~Krasser, and Y.~Tang, ``Mining dns for malicious domain
  registrations,'' in \emph{6th International Conference on Collaborative
  Computing: Networking, Applications and Worksharing (CollaborateCom
  2010)}.\hskip 1em plus 0.5em minus 0.4em\relax IEEE, 2010, pp. 1--6.

\bibitem{ref8}
J.~Ma, L.~K. Saul, S.~Savage, and G.~M. Voelker, ``Beyond blacklists: learning
  to detect malicious web sites from suspicious urls,'' in \emph{Proceedings of
  the 15th ACM SIGKDD international conference on Knowledge discovery and data
  mining}, 2009, pp. 1245--1254.

\bibitem{ref29}
M.~Antonakakis, R.~Perdisci, D.~Dagon, W.~Lee, and N.~Feamster, ``Building a
  dynamic reputation system for $\{$DNS$\}$,'' in \emph{19th USENIX Security
  Symposium (USENIX Security 10)}, 2010.

\bibitem{ref31}
W.~Liu, X.~Deng, G.~Huang, and A.~Y. Fu, ``An antiphishing strategy based on
  visual similarity assessment,'' \emph{IEEE Internet Computing}, vol.~10,
  no.~2, pp. 58--65, 2006.

\bibitem{ref34}
B.~Eshete, A.~Villafiorita, K.~Weldemariam, and M.~Zulkernine, ``Einspect:
  Evolution-guided analysis and detection of malicious web pages,'' in
  \emph{2013 IEEE 37th Annual Computer Software and Applications
  Conference}.\hskip 1em plus 0.5em minus 0.4em\relax IEEE Computer Society,
  2013, pp. 375--380.

\bibitem{ref12}
K.~Thomas, C.~Grier, J.~Ma, V.~Paxson, and D.~Song, ``Design and evaluation of
  a real-time url spam filtering service,'' in \emph{2011 IEEE symposium on
  security and privacy}.\hskip 1em plus 0.5em minus 0.4em\relax IEEE, 2011, pp.
  447--462.

\bibitem{ref13}
H.~Choi, B.~B. Zhu, and H.~Lee, ``Detecting malicious web links and identifying
  their attack types,'' in \emph{2nd USENIX Conference on Web Application
  Development (WebApps 11)}, 2011.

\bibitem{ref14}
G.~Xiang, J.~Hong, C.~P. Rose, and L.~Cranor, ``Cantina+ a feature-rich machine
  learning framework for detecting phishing web sites,'' \emph{ACM Transactions
  on Information and System Security (TISSEC)}, vol.~14, no.~2, pp. 1--28,
  2011.

\bibitem{ref15}
S.~Lee and J.~Kim, ``Warningbird: Detecting suspicious urls in twitter
  stream.'' in \emph{Ndss}, vol.~12, 2012, pp. 1--13.

\bibitem{ref50}
S.~N. Bannur, L.~K. Saul, and S.~Savage, ``Judging a site by its content:
  learning the textual, structural, and visual features of malicious web
  pages,'' in \emph{Proceedings of the 4th ACM Workshop on Security and
  Artificial Intelligence}, 2011, pp. 1--10.

\bibitem{ref51}
A.~Aggarwal, A.~Rajadesingan, and P.~Kumaraguru, ``Phishari: Automatic realtime
  phishing detection on twitter,'' in \emph{eCrime Researchers Summit}.\hskip
  1em plus 0.5em minus 0.4em\relax IEEE, 2012, pp. 1--12.

\bibitem{ref52}
H.~Choi, B.~B. Zhu, and H.~Lee, ``Detecting malicious web links and identifying
  their attack types,'' in \emph{2nd USENIX Conference on Web Application
  Development (WebApps 11)}, 2011.

\bibitem{ref16}
S.~Garera, N.~Provos, M.~Chew, and A.~D. Rubin, ``A framework for detection and
  measurement of phishing attacks,'' in \emph{Proceedings of the 2007 ACM
  workshop on Recurring malcode}, 2007, pp. 1--8.

\bibitem{ref17}
I.~Fette, N.~Sadeh, and A.~Tomasic, ``Learning to detect phishing emails,'' in
  \emph{Proceedings of the 16th international conference on World Wide Web},
  2007, pp. 649--656.

\bibitem{ref18}
A.~Bergholz, J.~H. Chang, G.~Paass, F.~Reichartz, and S.~Strobel, ``Improved
  phishing detection using model-based features.'' in \emph{CEAS}, 2008.

\bibitem{ref19}
S.~Abu-Nimeh, D.~Nappa, X.~Wang, and S.~Nair, ``A comparison of machine
  learning techniques for phishing detection,'' in \emph{Proceedings of the
  anti-phishing working groups 2nd annual eCrime researchers summit}, 2007, pp.
  60--69.

\bibitem{ref11}
H.~Chen, H.~Zhang, P.-Y. Chen, J.~Yi, and C.-J. Hsieh, ``Attacking visual
  language grounding with adversarial examples: A case study on neural image
  captioning,'' \emph{arXiv preprint arXiv:1712.02051}, 2017.

\bibitem{ref35}
M.~Barreno, B.~Nelson, R.~Sears, A.~D. Joseph, and J.~D. Tygar, ``Can machine
  learning be secure?'' in \emph{Proceedings of the ACM Symposium on
  Information, computer and communications security}, 2006, pp. 16--25.

\bibitem{ref24}
D.~Warde-Farley and I.~Goodfellow, ``11 adversarial perturbations of deep
  neural networks,'' \emph{Perturbations, Optimization, and Statistics}, vol.
  311, p.~5, 2016.

\bibitem{ref55}
P.-Y. Chen, H.~Zhang, Y.~Sharma, J.~Yi, and C.-J. Hsieh, ``Zoo: Zeroth order
  optimization based black-box attacks to deep neural networks without training
  substitute models,'' in \emph{Proceedings of the 10th ACM workshop on
  artificial intelligence and security}, 2017, pp. 15--26.

\end{thebibliography}

\bibliographystyle{IEEEtran}
\begin{IEEEbiography}[{\includegraphics[width=1in,height=1.2in]{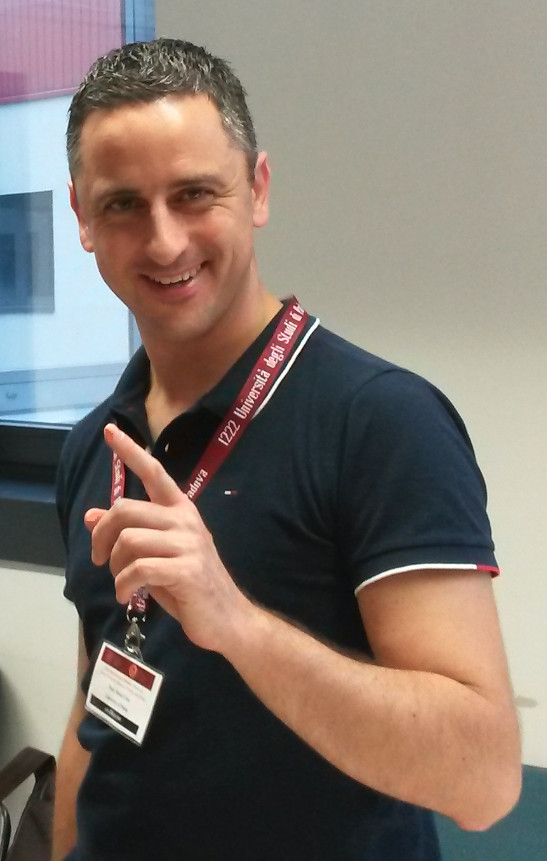}}]{Mauro Conti}
 is Full Professor at the University of Padua, Italy. He is also affiliated with TU Delft and University of Washington, Seattle. He obtained his Ph.D. from Sapienza University of Rome, Italy, in 2009. After his Ph.D., he was a Post-Doc Researcher at Vrije Universiteit Amsterdam, The Netherlands. In 2011 he joined as Assistant Professor at the University of Padua, where he became Associate Professor in 2015, and Full Professor in 2018. He has been Visiting Researcher at GMU, UCLA, UCI, TU Darmstadt, UF, and FIU. He has been awarded with a Marie Curie Fellowship (2012) by the European Commission, and with a Fellowship by the German DAAD (2013). His research is also funded by companies, including Cisco, Intel, and Huawei. His main research interest is in the area of Security and Privacy. In this area, he published more than 450 papers in topmost international peer-reviewed journals and conferences. He is Editor-in-Chief for IEEE Transactions on Information Forensics and Security, Area Editor-in-Chief for IEEE Communications Surveys \& Tutorials, and has been Associate Editor for several journals, including IEEE Communications Surveys \& Tutorials, IEEE Transactions on Dependable and Secure Computing, IEEE Transactions on Information Forensics and Security, and IEEE Transactions on Network and Service Management. He was Program Chair for TRUST 2015, ICISS 2016, WiSec 2017, ACNS 2020, CANS 2021, and General Chair for SecureComm 2012, SACMAT 2013, NSS 2021 and ACNS 2022. He is Fellow of the IEEE, Senior Member of the ACM, and Fellow of the Young Academy of Europe.
\end{IEEEbiography}
\vskip -1\baselineskip plus -1fil
\begin{IEEEbiography}[{\includegraphics[width=1in,height=1.2in]{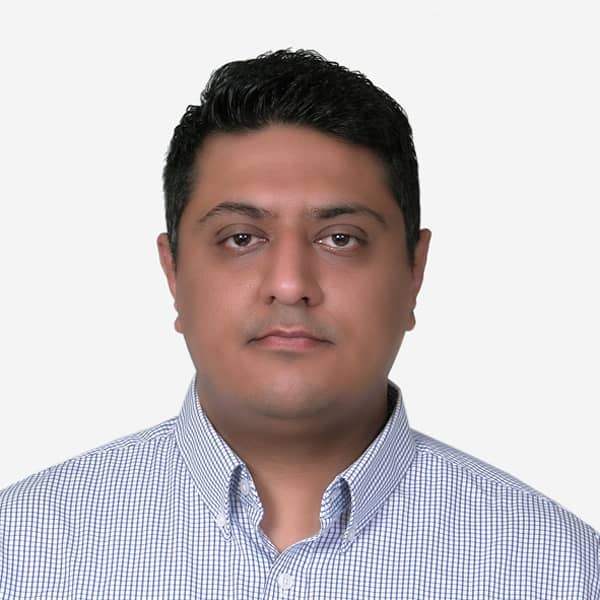}}]{Ehsan Nowroozi}
received his Ph.D. from Siena University in Italy. He is a research fellow at Sabanci University's Faculty of Engineering and Natural Sciences (FENS), Istanbul, Turkey 34956. Previously, he was a Postdoctoral Fellow at Siena and Padua Universities in Italy in 2020 and 2021, respectively. His research interests include security and privacy, with an emphasis on the use of image processing methods to multimedia authentication, network and Internet security. 
\end{IEEEbiography}
\vskip -2\baselineskip plus -1fil
\begin{IEEEbiography}[{\includegraphics[width=1in,height=1.2in]{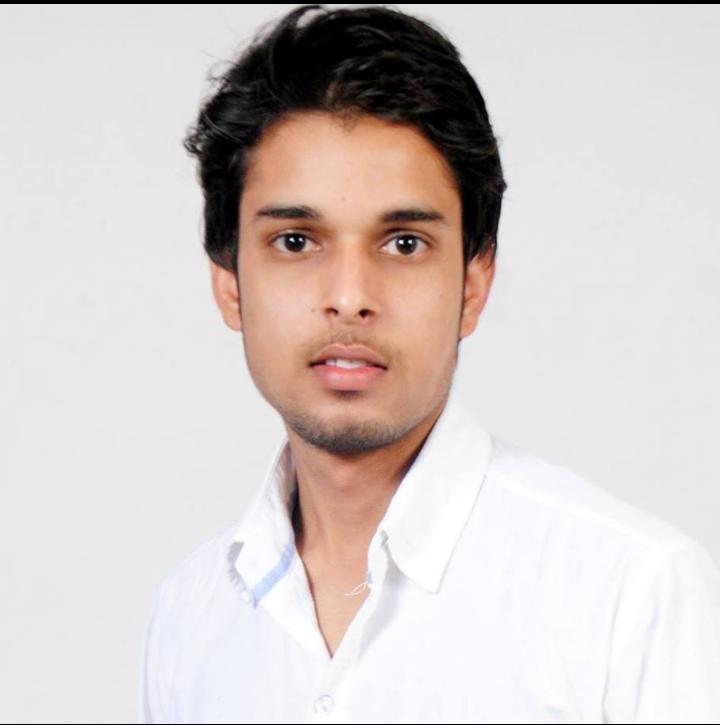}}]{Abhishek}
received the B.Sc. Degree in Honours in Mathematics from University of Delhi, India, M.Sc. Degree in Mathematics and Scientific Computing from MNNIT Allahabad, India. He worked in Artificial Intelligence and Digital Image Processing. Moreover, his research interest cover Machine Learning, Cryptography, Cyber-Security, Number Theory etc. Currently, he conducted his new research in applied machine learning in Cybersecurity with a research group from Italy. He is a member of IEEE Cybersecurity Community.
\end{IEEEbiography}
\vskip -2\baselineskip plus -1fil
\begin{IEEEbiography}[{\includegraphics[width=1in,height=1.2in]{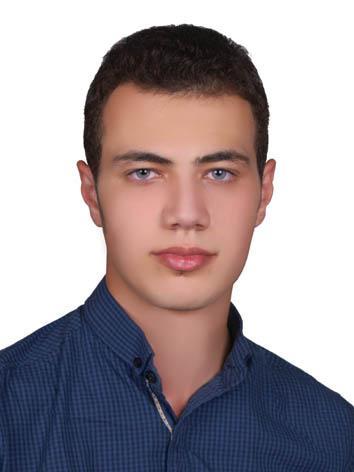}}]{Mohammadreza Mohammadi}
is a second-year Master student of ICT at University of Padua. He received the bachelor’s degree in computer engineering(software-network) from Bu-Ali Sina University, Hamedan, Iran in 2019. His main research interest is in the area of Machine Learning, Cybersecurity, IoT and Computer Vision. He worked in Industrial IoT security and AI, and Intrusion detection systems(IDS) and Healthcare systems and He is also a graduate student member of IEEE institution.\end{IEEEbiography}

\end{document}